# Saving Gradient and Negative Curvature Computations: Finding Local Minima More Efficiently

Yaodong Yu[*][‡]    and    Difan Zou[†][‡]    and    Quanquan Gu[§]

## Abstract

We propose a family of nonconvex optimization algorithms that are able to save gradient and negative curvature computations to a large extent, and are guaranteed to find an approximate local minimum with improved runtime complexity. At the core of our algorithms is the division of the entire domain of the objective function into small and large gradient regions: our algorithms only perform gradient descent based procedure in the large gradient region, and only perform negative curvature descent in the small gradient region. Our novel analysis shows that the proposed algorithms can escape the small gradient region in only one negative curvature descent step whenever they enter it, and thus they only need to perform at most $N_\epsilon$ negative curvature direction computations, where $N_\epsilon$ is the number of times the algorithms enter small gradient regions. For both deterministic and stochastic settings, we show that the proposed algorithms can potentially beat the state-of-the-art local minima finding algorithms. For the finite-sum setting, our algorithm can also outperform the best algorithm in a certain regime.

## 1 Introduction

We consider the following unconstrained optimization problem

$$\min_{\mathbf{x}\in\mathbb{R}^d} f(\mathbf{x}), \tag{1.1}$$

where $f: \mathbb{R}^d \to \mathbb{R}$ is twice differentiable and can be nonconvex. Note that the problem in (1.1) becomes a stochastic optimization if $f$ is the expectation of some underlying function indexed by some random variable, or a finite-sum optimization if $f$ is the average of many component functions. For general nonconvex optimization problems, finding a global minimum can be NP-hard (Hillar and Lim, 2013). Therefore, instead of finding the global minimum, a more modest goal is to find a local minimum of (1.1). In fact, for many nonconvex optimization problems in machine learning, it is sufficient to find a local minimum. Previous work in deep neural networks (Choromanska et al., 2015; Dauphin et al., 2014) showed that a local minimum can be as good as a global minimum. Moreover, recent work showed that for some machine learning problems such as matrix completion (Ge et al., 2016), matrix sensing (Bhojanapalli et al., 2016; Park et al., 2016) and phase retrieval (Sun et al., 2016), there is no spurious local minimum, i.e., all local minima are global minima. This also justifies the design of algorithms that guarantee convergence to a local minimum.

---

[*]Department of Computer Science, University of Virginia, Charlottesville, VA 22904, USA; e-mail:`yy8ms@virginia.edu`

[†]Department of Systems and Information Engineering, University of Virginia, Charlottesville, VA 22904, USA; e-mail: `dz5an@virginia.edu`

[‡]Equal Contribution

[§]Department of Computer Science, University of Virginia, Charlottesville, VA 22904, USA; e-mail: `qg5w@virginia.edu`

More specifically, we aim to find an approximate local minimum of (1.1), which satisfies the following $(\epsilon, \epsilon_H)$-second-order necessary optimality condition,

$$\|\nabla f(\mathbf{x})\|_2 \leq \epsilon, \text{ and } \lambda_{\min}(\nabla^2 f(\mathbf{x})) \geq -\epsilon_H, \tag{1.2}$$

where $\epsilon, \epsilon_H \in (0, 1)$ are predefined tolerance parameters. For $\rho$-Hessian Lipschitz functions, if we choose $\epsilon_H = \sqrt{\rho\epsilon}$, (1.2) reduces to the definition of $\epsilon$-second-order stationary point in Nesterov and Polyak (2006). In the sequel, we will use approximate local minimum and approximate second-order stationary point interchangeably.

Cubic regularized Newton's method (Nesterov and Polyak, 2006), which is based on the second-order Hessian information of the function, is arguably the first provable method that can find the approximate local minima (i.e., approximate second-order stationary points). However, the runtime complexity of cubic regularization is very high, because it needs to solve a very expensive cubic problem in each iteration. In order to overcome this computational burden, Agarwal et al. (2016) proposed to solve the cubic problem using approximate matrix inversion. In a concurrent work, Carmon and Duchi (2016) proposed to use gradient descent to solve the cubic problem. Both of these methods still need to solve the cubic problem approximately in each iteration. In another line of research, instead of using the cubic regularization technique, negative curvature direction is directly computed for finding local minima. For example, by exploiting the structure of almost convexity, Carmon et al. (2016) proposed a nonconvex optimization method based on accelerated gradient descent and negative curvature descent, which only needs gradient and Hessian-vector evaluations. Similar idea has been used in Allen-Zhu (2017) for stochastic nonconvex optimization. Nonetheless, both algorithms need to calculate the negative curvature direction in every iteration, which makes them less practical. On the other hand, first-order methods with random noise injection (Ge et al., 2015; Levy, 2016; Jin et al., 2017a) have been shown to be able to find the approximate second-order stationary points as well. Yet the runtime complexity of these first-order algorithms often has a worse dependence on $\epsilon$ than the aforementioned second-order algorithms. Very recently, Xu and Yang (2017); Allen-Zhu and Li (2017b) showed that noise injection idea in previous work (Jin et al., 2017a) is essentially a way to find the negative curvature direction. Based on this insight, they proposed first-order methods which only need to access first-order oracles and are able to find local minima as fast as the state-of-the-art Hessian-vector product-based algorithms (Agarwal et al., 2016; Carmon et al., 2016; Allen-Zhu, 2017). Nevertheless, these algorithms (Xu and Yang, 2017; Allen-Zhu and Li, 2017b) need to perform gradient descent or its variations in each iteration, even the algorithm is close to a first-order stationary point. Therefore, one interesting question is: in order to find local minima more efficiently, can we design a practical algorithm that can save gradient and negative curvature computation as much as possible?

In this paper, we show that the answer to the above question is affirmative. We develop a family of algorithms that are able to save the gradient and negative curvature computation to a large extent. The high-level idea is: we divide the entire domain of the objective function into two regions based on the magnitude of the gradient norm, i.e., $\|\nabla f(\mathbf{x})\|_2$. We define the large gradient region as $\{\mathbf{x} : \|\nabla f(\mathbf{x})\|_2 > \epsilon\}$ and define the small gradient region as $\{\mathbf{x} : \|\nabla f(\mathbf{x})\|_2 \leq \epsilon\}$. The proposed algorithms will only perform gradient descent-based methods in the large gradient region, and only perform negative curvature descent in the small gradient region. Furthermore, we will show that a single negative curvature descent step is sufficient for our algorithms to escape from the small gradient region, and therefore our algorithms must perform gradient descent-based methods in the next iteration. This dramatically saves the gradient (or stochastic gradient) and negative curvature computations, and improves the overall runtime complexity.

**Our Contributions** We summarize our major contributions as follows:

- We propose a family of practical algorithms, which are guaranteed to find an approximate local minimum with improved runtime complexity. The proposed algorithms are able to save gradient and negative curvature computation strikingly. Our novel analysis shows that the proposed algorithms can escape from each saddle point in one negative curvature descent step, and therefore only need to compute negative curvature directions at most $N_\epsilon$ times, where $N_\epsilon$ is the number of times the algorithms enter small gradient regions. Thus the runtime can potentially be much less than that of existing state-of-the-art algorithms.

- Our proposed algorithms for finding local minima cover deterministic, stochastic and finite-sum settings. For both deterministic and stochastic settings, we show that the proposed algorithms can potentially outperform the state-of-the-art approximate local minima finding algorithms in terms of runtime complexity. For the finite-sum setting, our algorithm can be better than the state-of-the-art algorithms in a certain regime.

- Our novel proof technique is based on a refined analysis of the runtime complexity, which counts the number of gradient descent steps and the number of negative curvature descent steps in a unified way, and rigorously integrates the high-probability and expectation-based arguments. We believe that our proof technique is of independent interest and can be applied to improve the analyses of many existing nonconvex optimization algorithms.

The remainder of this paper is organized as follows: We review and discuss the related work in Section 2, and introduce some definitions in Section 3. We present our key idea, algorithm and theoretical analysis for deterministic nonconvex optimization in Section 4. In Section 5, we present an algorithm and theoretical analysis in the stochastic optimization setting. In Section 6, we present an algorithm and theoretical analysis for finite-sum nonconvex optimization. Finally, we conclude our paper and point out some future directions in Section 7.

**Notation:** Let $\mathbf{A} = [A_{ij}] \in \mathbb{R}^{d \times d}$ be a matrix and $\mathbf{x} = (x_1, ..., x_d)^\top \in \mathbb{R}^d$ be a vector. We use $\|\mathbf{x}\|_q = (\sum_{i=1}^d |x_i|^q)^{1/q}$ to denote $\ell_q$ vector norm for $0 < q < +\infty$. Denote the spectral and Frobenius norm of $\mathbf{A}$ by $\|\mathbf{A}\|_2$ and $\|\mathbf{A}\|_F$. For a symmetric matrix $\mathbf{A}$, we denote by $\lambda_{\max}(\mathbf{A})$, $\lambda_{\min}(\mathbf{A})$ and $\lambda_i(\mathbf{A})$ the maximum, minimum and $i$-th largest eigenvalues of $\mathbf{A}$. We denote by $\mathbf{A} \succeq 0$ that $\mathbf{A}$ is positive semidefinite (PSD). Given two sequences $\{a_n\}$ and $\{b_n\}$, we write $a_n = O(b_n)$ if there exists a constant $0 < C < +\infty$ such that $a_n \leq C\, b_n$. We use notation $\widetilde{O}(\cdot)$ to hide the logarithmic factors. We also make use of the notation $f_n \lesssim g_n$ ($f_n \gtrsim g_n$) if $f_n$ is less than (larger than) $g_n$ up to a constant.

# 2 Related Work

For nonconvex optimization problems, there is a vast literature on algorithms for finding approximate local minima (i.e., approximate second-order stationary points). In general, existing work can be divided into three setting: deterministic optimization, stochastic optimization and finite-sum optimization. We will discuss the related work in each setting respectively.

## 2.1 Deterministic Setting

In the deterministic setting, algorithms can access to the full gradient and Hessian information. Originally proposed by Nesterov and Polyak (2006), the cubic regularization algorithm converges to the $(\epsilon, \epsilon_H)$-second-order stationary point in $O\big(\max\{\epsilon^{-3/2}, \epsilon_H^{-3}\}\big)$ iterations. Cartis et al. (2012) showed that the classical trust region Newton method can find the $(\epsilon, \epsilon_H)$-second-order stationary point after at most $O\big(\max\{\epsilon^{-2}\epsilon_H^{-1}, \epsilon_H^{-3}\}\big)$ iterations. Later the iteration complexity of trust region Newton methods is improved to be the same as cubic regularization (Curtis et al., 2017; Martínez and Raydan, 2017). Yet the cubic regularization method and trust region Newton method suffer from solving a very expensive subproblem in each iteration. In order to alleviate the per-iteration cost of cubic regularization, several recent work (Agarwal et al., 2016; Carmon and Duchi, 2016) was proposed to solve the cubic subproblem approximately and achieve better runtime complexity. While these variants of cubic regularization/trust region Newton based methods only need to compute Hessian-vector product, they still need to access the Hessian information in each iteration.

In addition to cubic regularization and trust region Newton methods, utilizing negative curvature directions is also able to find second-order stationary points for nonconvex problems, which dates back to McCormick (1977); Moré and Sorensen (1979); Goldfarb (1980), and is adopted in some recent work (Carmon et al., 2016; Royer and Wright, 2017) as well. In detail, Carmon et al. (2016) proposed to use the accelerated gradient method together with the negative curvature direction to find an approximate second-order stationary point, which requires $\widetilde{O}(\epsilon^{-7/4})$ gradient and Hessian-vector product evaluation if choosing $\epsilon_H = O(\sqrt{\epsilon})$. However, it needs to perform the negative curvature descent step before taking accelerated gradient descent in each iteration. Royer and Wright (2017) proposed an algorithm mainly based on line search, which involves computing the negative curvature direction and Newton direction in each iteration. Its iteration complexity also matches the best-known iteration complexity of second-order algorithms.

Different from the above approaches, there is another line of research (Levy, 2016; Jin et al., 2017a; Xu and Yang, 2017; Allen-Zhu and Li, 2017b; Jin et al., 2017b), which shows that without second-order information, it is possible to find a second-order stationary point using first order information with random noise injection. The best-known runtime complexity of this kind of methods in the deterministic setting is $O\big(\log^6(d)\,\epsilon^{-7/4}\big)$ if setting $\epsilon_H = \sqrt{\epsilon}$, which is presented in a very recent work (Jin et al., 2017b).

## 2.2 Stochastic Setting

In the stochastic setting, the algorithms cannot access the full gradient and Hessian directly. Instead, they can only access the stochastic gradient and stochastic Hessian. Ge et al. (2015) analyzed a variant of stochastic gradient descent (SGD) method by adding noise for nonconvex optimization, and showed that its computational complexity for finding $(\epsilon, \sqrt{\epsilon})$-second-order stationary points is $O(\epsilon^{-4}\text{poly}(d))$. Based on the mechanism of variance reduction and negative curvature descent, Allen-Zhu (2017) proposed an algorithm which is able to find second-order stationary points faster than SGD, and its runtime complexity is $\widetilde{O}(\epsilon^{-7/2})$ if $\epsilon_H = \sqrt{\epsilon}$. Similar to Carmon et al. (2016), it also needs to perform negative curvature descent before applying stochastic gradient descent based methods in each outer iteration. Kohler and Lucchi (2017); Xu et al. (2017) proposed to use subsampled Hessian to replace the full Hessian in cubic regularization and/or trust region method, and provided certain conditions that enable the use of subsampled Hessian while still attain the iteration complexity of the original exact methods. Tripuraneni et al. (2017) designed a stochastic algorithm based on cubic regularization and used $\widetilde{O}(\epsilon^{-7/2})$ stochastic gradient and Hessian-vector product evaluations to find $(\epsilon, \sqrt{\epsilon})$-second-order stationary points. Very recently, Xu and Yang (2017); Allen-Zhu and Li (2017b) turned the stochastically controlled stochastic gradient (SCSG) method (Lei et al., 2017) into a local-minimum finding algorithm, by incorporating a first-order negative curvature finding procedure. The computational complexity of algorithms in Xu and Yang (2017); Allen-Zhu and Li (2017b) is $\widetilde{O}(\epsilon^{-10/3} + \epsilon^{-2}\epsilon_H^{-3})$ for finding $(\epsilon, \epsilon_H)$-second-order stationary points. However, the algorithms in Xu and Yang (2017) need to compute the negative curvature in each iteration, and the algorithms in Allen-Zhu and Li (2017b) need to perform gradient descent based methods in each iteration, and sometimes followed by a negative curvature descent in the same iteration.

### 2.3 Finite-Sum Setting

The finite-sum optimization motivated the development of variance reduction based methods (Johnson and Zhang, 2013; Reddi et al., 2016; Allen-Zhu and Hazan, 2016). In this setting, Agarwal et al. (2016) proposed an algorithm to find $(\epsilon, \sqrt{\epsilon})$-second-order stationary points with $\widetilde{O}(n\epsilon^{-3/2} + n^{3/4}\epsilon^{-7/4})$ gradient and Hessian-vector product evaluations. Reddi et al. (2017) introduced a generic framework to find $(\epsilon, \epsilon_H)$-second-order stationary points, and its overall runtime complexity is $\widetilde{O}(n^{2/3}\epsilon^{-2} + n\epsilon_H^{-3} + n^{3/4}\epsilon_H^{-7/2})$ by using gradient and Hessian-vector product evaluations. Very recently, Allen-Zhu and Li (2017b) turned the nonconvex stochastic variance reduced gradient (SVRG) (Lei et al., 2017) into a local-minimum finding algorithm, using the first-order negative curvature finding procedure proposed in Xu and Yang (2017); Allen-Zhu and Li (2017b). The runtime complexity of the resulting algorithm is $\widetilde{O}(n^{2/3}\epsilon^{-2} + n\epsilon_H^{-3} + n^{3/4}\epsilon_H^{-7/2} + n^{5/12}\epsilon^{-2}\epsilon_H^{-1/2})$. Similar to the stochastic setting, the drawback of this algorithm is that it may need to perform SVRG and negative curvature descent in the same iteration.

To compare our algorithms with the state-of-the-art methods in a more comprehensive way, we summarize the runtime complexity of the state-of-the-art methods and our methods in Table 1. Here we keep the second-order accuracy parameter $\epsilon_H$ for some algorithms for the ease of comparison, and we can set $\epsilon_H = O(\sqrt{\epsilon})$ for different algorithms in Table 1 to obtain the simplified results. Note that we only show the runtime complexity of our algorithms when using the first-order negative curvature finding procedure (Allen-Zhu and Li, 2017b), and omit the results of our algorithms when using Hessian-vector product based negative finding curvature procedures (Kuczyński and Woźniakowski, 1992; Garber et al., 2016; Allen-Zhu and Li, 2017a). Note also that the algorithms in Carmon et al. (2016); Allen-Zhu (2017); Reddi et al. (2017) can be converted to Hessian-vector product free methods by replacing the fast PCA based negative curvature finding procedure with the first-order negative curvature finding procedure (Xu and Yang, 2017; Allen-Zhu and Li, 2017b). As we can see from Table 1, in the deterministic and stochastic settings, the runtime complexity of our algorithms can be better than other algorithms, if the number of times that our algorithms enter small gradient regions, i.e., $N_\epsilon$, is substantially smaller than $O(\epsilon_H^{-3})$. And in the worst case, the runtime complexity of our algorithms match the best results in these two settings. In the finite-sum setting, when $n \gtrsim \epsilon^{-3/2}$, our algorithm outperforms the other algorithms if $N_\epsilon$ is substantially smaller and matches the best runtime result in the worst case.

Table 1: A comparison of different methods that are guaranteed to converge to second-order stationary point in terms of run-time complexity. Here $N_\epsilon$ is the number of times our algorithms enter the small gradient region. We let $\mathbb{T}_g$ be the time complexity of gradient (or stochastic gradient) evaluation, and $\mathbb{T}_h$ be the time complexity of Hessian-vector product (or stochastic Hessian-vector product) evaluation. Following the convention of literature, $\mathbb{T}_g$ and $\mathbb{T}_h$ are considered in the same order and hence we omit them in the big-$\widetilde{O}$ notation.

| | Algorithm | Runtime Complexity | Hessian-Vector Product |
|---|---|---|---|
| Deterministic | ANCM (Carmon et al., 2016) | $\widetilde{O}\left(\frac{1}{\epsilon^{7/4}} + \frac{1}{\epsilon_H^{7/2}}\right)$ | needed |
| | FastCubic (Agarwal et al., 2016) | $\widetilde{O}\left(\frac{1}{\epsilon^{3/2}} + \frac{1}{\epsilon^{7/4}}\right)$ | needed |
| | PAGD (Jin et al., 2017b) | $\widetilde{O}\left(\frac{1}{\epsilon^{7/4}}\right)$ | not needed |
| | GOSE$^{\text{det}}$ | $\widetilde{O}\left(\frac{1}{\epsilon^{7/4}} + \left(\frac{1}{\epsilon^{1/4}} + \frac{1}{\epsilon_H^{1/2}}\right)\min\left\{\frac{1}{\epsilon_H^3}, N_\epsilon\right\}\right)$ | not needed |
| Stochastic | Natasha2 (Allen-Zhu, 2017) | $\widetilde{O}\left(\frac{1}{\epsilon^3 \epsilon_H} + \frac{1}{\epsilon^{13/4}} + \frac{1}{\epsilon_H^5}\right)$ | needed |
| | StochasticCubic (Tripuraneni et al., 2017) | $\widetilde{O}\left(\frac{1}{\epsilon^{7/2}}\right)$ | needed |
| | Neon2+SCSG (Allen-Zhu and Li, 2017b) | $\widetilde{O}\left(\frac{1}{\epsilon^{10/3}} + \frac{1}{\epsilon^2 \epsilon_H^3} + \frac{1}{\epsilon_H^5}\right)$ | not needed |
| | GOSE$^{\text{stochastic}}$ | $\widetilde{O}\left(\frac{1}{\epsilon^{10/3}} + \frac{1}{\epsilon^2 \epsilon_H^3} + \left(\frac{1}{\epsilon^2} + \frac{1}{\epsilon_H^2}\right)\min\left\{\frac{1}{\epsilon_H^3}, N_\epsilon\right\}\right)$ | not needed |
| Finite-Sum | FastCubic (Agarwal et al., 2016) | $\widetilde{O}\left(\frac{n}{\epsilon^{3/2}} + \frac{n^{3/4}}{\epsilon^{7/4}}\right)$ | needed |
| | Mix (Reddi et al., 2017) | $\widetilde{O}\left(\frac{n^{2/3}}{\epsilon^2} + \frac{n}{\epsilon_H^3} + \frac{n^{3/4}}{\epsilon_H^{7/2}}\right)$ | needed |
| | Neon2+SVRG (Allen-Zhu and Li, 2017b) | $\widetilde{O}\left(\frac{n^{2/3}}{\epsilon^2} + \frac{n}{\epsilon_H^3} + \frac{n^{3/4}}{\epsilon_H^{7/2}} + \frac{n^{5/12}}{\epsilon^2 \epsilon_H^{1/2}}\right)$ | not needed |
| | GOSE$^{\text{finite-sum}}$ | $\widetilde{O}\left(\frac{n^{2/3}}{\epsilon^2} + \left(n + \frac{n^{3/4}}{\epsilon_H^{1/2}}\right)\min\left\{\frac{1}{\epsilon_H^3}, N_\epsilon\right\}\right)$ | not needed |

## 3 Preliminary Definitions

In this section, we will introduce some definitions which will be used later.

**Definition 3.1** (Gradient Lipschitz). A differentiable function $f(\cdot)$ is $L$-gradient Lipschitz, if for any $\mathbf{x}, \mathbf{y} \in \mathbb{R}^d$, we have
$$\|\nabla f(\mathbf{x}) - \nabla f(\mathbf{y})\|_2 \leq L \|\mathbf{x} - \mathbf{y}\|_2.$$

Typically, $L$-gradient Lipschitz implies that the eigenvalues of Hessian $\nabla^2 f(\cdot)$ are confined in the region $[-L, L]$, and this property is also known as $L$-smoothness.

**Definition 3.2** (Hessian Lipschitz). A twice-differentiable function $f(\cdot)$ is $\rho$-Hessian Lipschitz, if for any $\mathbf{x}, \mathbf{y} \in \mathbb{R}^d$, we have
$$\|\nabla^2 f(\mathbf{x}) - \nabla^2 f(\mathbf{y})\|_2 \leq \rho \|\mathbf{x} - \mathbf{y}\|_2.$$

**Definition 3.3** (Strongly Convex). A twice-differentiable function $f(\cdot)$ is $\mu$-strongly convex, if for any $\mathbf{x} \in \mathbb{R}^d$, we have
$$\lambda_{\min}(\nabla^2 f(\mathbf{x})) \geq \mu.$$

**Definition 3.4** (Optimal Gap). For a function $f(\cdot)$, we introduce $\Delta_f$ at point $\mathbf{x}_0$ which is defined as
$$f(\mathbf{x}_0) - \inf_{\mathbf{y} \in \mathbb{R}^d} f(\mathbf{y}) \leq \Delta_f,$$

without loss of generality, we assume $\Delta_f < +\infty$.

**Definition 3.5** (Geometric Distribution). For a random integer $N$, we say it satisfies the geometric distribution with parameter $p$, denoted as $\text{Geom}(p)$, if it follows that
$$\Pr(N = k) = p^k(1 - p), \quad \forall k = 0, 1, \ldots.$$

We let $\mathbb{T}_g$ be the time complexity of gradient (or stochastic gradient) evaluation, and $\mathbb{T}_h$ be time complexity of Hessian-vector product (stochastic Hessian-vector product) evaluation. For the ease of presentation, and also following the convention of the literature, we consider $\mathbb{T}_g$ and $\mathbb{T}_h$ to be in the same order. Thus, we use $\mathbb{T}_g$ to denote both $\mathbb{T}_g$ and $\mathbb{T}_h$ in the sequel.

## 4 Deterministic Nonconvex Optimization

In this section, we present our proposed algorithm for deterministic nonconvex optimization problem in (1.1). We first introduce the key idea of how to save the gradient and negative curvature evaluations and how to escape each saddle point in one step, and then present our algorithm and a novel runtime analysis.

### 4.1 Key Idea

To avoid frequent negative curvature direction computation, we first divide the domain of the objective function into two different regions based on the magnitude of the gradient norm: **large gradient region** ($\|\nabla f(\mathbf{x})\|_2 > \epsilon$) and **small gradient region** ($\|\nabla f(\mathbf{x})\|_2 \leq \epsilon$). This enables us to apply gradient-based and negative curvature-based methods respectively in these two regions.
**Large gradient region:** When the norm of the gradient is large, which means that the iterate $\mathbf{x}_k$ is in a large gradient region, we apply gradient descent based methods to make the objective function decrease. For instance, based on the well-known convergence result of gradient descent in Nesterov (1998), when the function $f(\cdot)$ is $L$-gradient Lipschitz, it can always decrease the objective function value when the step size $\eta \leq 1/L$.

**Small gradient region:** When the iterate $\mathbf{x}_k$ enters the small gradient region, second-order information is taken into account to escape from this region if the iterate is not an approximate local minimum. In detail, suppose $\mathbf{x}_k$ satisfies

$$\|\nabla f(\mathbf{x}_k)\|_2 \leq \epsilon, \text{ and } \lambda_{\min}(\nabla^2 f(\mathbf{x}_k)) < -\epsilon_H, \tag{4.1}$$

which suggests that there exists a negative curvature direction around $\mathbf{x}_k$. Otherwise, $\mathbf{x}_k$ is already the $(\epsilon, \epsilon_H)$-second-order stationary point. Note that the small gradient region with negative curvature can be regarded as saddle point region. And it will become smaller as $\epsilon$ goes to 0, which indicates that it is easy to escape from the saddle point region given an appropriate direction. Following many existing algorithms (Carmon et al., 2016; Allen-Zhu, 2017; Reddi et al., 2017), we adopt negative curvature descent (McCormick, 1977; Moré and Sorensen, 1979; Goldfarb, 1980) to escape from the saddle point region.

More specifically, suppose $\lambda_{\min}(\nabla^2 f(\mathbf{x}_k)) < -\epsilon_H$, a direction $\widehat{\mathbf{v}} \in \mathbb{R}^d$ is called a negative curvature direction if it satisfies

$$\widehat{\mathbf{v}}^\top \nabla^2 f(\mathbf{x}_k) \widehat{\mathbf{v}} < -\frac{\epsilon_H}{2}.$$

The negative curvature descent is as follows

$$\mathbf{x}_{k+1} = \mathbf{x}_k + \eta \widetilde{\mathbf{v}}, \tag{4.2}$$

where $\widetilde{\mathbf{v}} = \text{sign}(-\nabla f(\mathbf{x}_k)^\top \widehat{\mathbf{v}}) \widehat{\mathbf{v}}$ is the adjusted negative curvature direction, and $\eta > 0$ is the step size parameter which will be specified later. In order to reduce negative curvature computation, we try to perform negative curvature step as few as possible. We will show that by appropriately choosing the step size $\eta$, the negative curvature descent step enjoys the following two properties, which are formally stated later in this section,

- After taking a negative curvature descent step, the objective function value will decrease, i.e., $f(\mathbf{x}_{k+1}) - f(\mathbf{x}_k) < -O(\epsilon_H^3)$.

- Gradient norm will increase after a negative curvature descent step, i.e., $\|\nabla f(\mathbf{x}_{k+1})\|_2 > \|\nabla f(\mathbf{x}_k)\|_2$, and $\|\nabla f(\mathbf{x}_{k+1})\|_2 > \epsilon$.

These two properties ensure that the negative curvature descent step is able to decrease the function value and escape from the small gradient region $\|\nabla f(\mathbf{x})\| \leq \epsilon$ at the same time. And the algorithm will resume a gradient descent based method until it gets into another small gradient region.

## 4.2 Algorithm

Now we present our algorithm for finding local minima in the deterministic setting. The primary way to find a negative curvature direction is by computing the eigenvector corresponding to the smallest eigenvalue of the Hessian matrix $\nabla^2 f(\mathbf{x}_k)$, which incurs at least $O(d^2)$ computational complexity. Recently, some efficient approaches (Kuczyński and Woźniakowski, 1992; Garber et al., 2016; Xu and Yang, 2017; Allen-Zhu and Li, 2017b) have been proposed and investigated in order to find the negative curvature direction more efficiently.

One type of methods is to compute the leading eigenvector of the shift-and-invert Hessian matrix approximately using existing fast eigenvalue decomposition/singular value decomposition algorithms. For example, we can use the Lanczos method (Kuczyński and Woźniakowski, 1992) and the fast PCA method (Garber et al., 2016). In the sequel, we refer to this type of methods as FastPCA. By using any of these FastPCA algorithms, we can obtain an $\epsilon_H$-approximate eigenvector with probability at least $1 - \delta$ in a small number of matrix-vector product operations, as spelled out in the following lemma.

**Lemma 4.1** (Kuczyński and Woźniakowski (1992)). Let $f(\cdot)$ be a function that is $L$-smooth and $\rho$-Hessian Lipschitz continuous. For any $\mathbf{x} \in \mathbb{R}^d$, let $\mathbf{H} = \nabla^2 f(\mathbf{x})$. If $\lambda_{\min}(\mathbf{H}) < -\epsilon_H$, then with probability at least $1 - \delta$, Lanczos method will return a unit vector $\widehat{\mathbf{v}}$ satisfying

$$\widehat{\mathbf{v}}^\top \mathbf{H} \widehat{\mathbf{v}} < -\frac{\epsilon_H}{2},$$

in at most $O\big(\log(d/\delta)\sqrt{L/\epsilon_H}\big)$ Hessian-vector product evaluations.

Recently, another type of approaches (Xu and Yang, 2017; Allen-Zhu and Li, 2017b) has been proposed to compute the negative curvature direction without Hessian-vector product computation. By adding some random perturbation at the beginning, and updating the iterates based on first-order information, they are able to extract the approximate negative curvature. One of these algorithms is called Neon2 (Allen-Zhu and Li, 2017b) and we summarize its result in the following lemma.

**Lemma 4.2** (Allen-Zhu and Li (2017b)). Let $f(\cdot)$ be a function that is $L$-smooth and $\rho$-Hessian Lipschitz continuous. For any $\mathbf{x} \in \mathbb{R}^d$, let $\mathbf{H} = \nabla^2 f(\mathbf{x})$. With probability at least $1 - \delta$, Neon2$^{\text{det}}$ returns $\widehat{\mathbf{v}}$ satisfying one of the following conditions,

- $\widehat{\mathbf{v}} = \bot$, then $\lambda_{\min}(\mathbf{H}) \geq -\epsilon_H$.
- $\widehat{\mathbf{v}} \neq \bot$, then $\widehat{\mathbf{v}}^\top \mathbf{H} \widehat{\mathbf{v}} \leq -\epsilon_H/2$ with $\|\widehat{\mathbf{v}}\|_2 = 1$.

The total number of gradient evaluations is $O\big(\log^2(d/\delta)\sqrt{L/\epsilon_H}\big)$.

Note that in both Lemmas 4.1 and 4.2, the iteration complexity depends on $\log d$ factor. This is because the logarithmic dependence on the dimension $d$ in eigenvector approximation is unavoidable (Simchowitz et al., 2017).

To summarize, there exists an algorithm, denoted by ApproxNC$(\cdot)$, that returns a unit vector $\widehat{\mathbf{v}}$ such that $\widehat{\mathbf{v}}^\top \mathbf{H} \widehat{\mathbf{v}} \leq -\epsilon_H/2$ if $\lambda_{\min}(\nabla^2 f(\mathbf{x})) < -\epsilon_H$, otherwise it returns $\widehat{\mathbf{v}} = \bot$. Next we present our Algorithm 1, which is able to escape from a saddle point in one step.

---

**Algorithm 1** One-Step-Deterministic $(f(\cdot), \mathbf{x}, \epsilon_H, \rho, \delta)$

---
1: Set $\eta \leftarrow \epsilon_H/(2c_1\rho)$
2: $\widehat{\mathbf{v}} \leftarrow \text{ApproxNC}\,(f(\cdot), \mathbf{x}, \epsilon_H, \delta)$
3: **if** $\widehat{\mathbf{v}} \neq \bot$ **then**
4: $\quad \widetilde{\mathbf{v}} \leftarrow \text{sign}(-\nabla f(\mathbf{x})^\top \widehat{\mathbf{v}})\widehat{\mathbf{v}}$
5: $\quad \mathbf{y} = \mathbf{x} + \eta \widetilde{\mathbf{v}}$
6: $\quad$ **return** $\mathbf{y}$
7: **else**
8: $\quad$ **return** $\bot$

---

As described in Algorithm 1, we first estimate the negative curvature of function $f(\cdot)$ at $\mathbf{x}$, by using FastPCA (Garber et al., 2016) or Neon2 (Allen-Zhu and Li, 2017b). If $\widehat{\mathbf{v}} \neq \perp$, we can adjust the direction of $\widehat{\mathbf{v}}$ as $\widetilde{\mathbf{v}}$ according to the gradient $\nabla f(\mathbf{x})$, take a negative curvature descent step along $\widetilde{\mathbf{v}}$ with appropriate step size $\eta$ and return $\mathbf{y}$. Otherwise Algorithm 1 will return $\perp$. As we will prove later in this section, one can use Algorithm 1 to escape a saddle point $\mathbf{x}$ with $\lambda_{\min}(\nabla^2 f(\mathbf{x})) < -\epsilon_H$.

Next we are going to present a practical algorithm for finding local minima in the deterministic setting, which is summarized in Algorithm 2.

---

**Algorithm 2** **G**radient descent with **O**ne-**S**tep **E**scaping, GOSE-Deterministic $(\mathbf{x}_0, \epsilon, \epsilon_H, L, \rho, \delta)$

1: **for** $k = 1, 2, ...$ **do**
2:   **if** $\|\nabla f(\mathbf{x}_{k-1})\|_2 > \epsilon$
3:     $\mathbf{x}_k \leftarrow \text{GN-AGD}(f(\cdot), \mathbf{x}_{k-1}, L, \epsilon, \epsilon^{1/2}, \epsilon^{1/2}/\rho)$
4:   **else**
5:     $\mathbf{x}_k \leftarrow \text{One-Step-Deterministic}(f(\cdot), \mathbf{x}_{k-1}, \epsilon_H, \rho, \delta)$
6:     **if** $\mathbf{x}_k = \perp$
7:       **return** $\mathbf{x}_{k-1}$
8: **endfor**

---

As we can see from the pseudocode in Algorithm 2, for large gradient regions, we apply the Guarded Nonconvex Accelerated Gradient Descent algorithm (GN-AGD($\cdot$)) (Carmon et al., 2017), a variant of accelerated gradient descent method for minimizing nonconvex functions, to find $\epsilon$-first-order stationary points. Please refer to Algorithm 3 in Carmon et al. (2017) for more details. According to the analysis in Carmon et al. (2017), GN-AGD($\cdot$) can find $\epsilon$-first-order stationary points faster than gradient descent. More specifically, to find $\epsilon$-first-order stationary points, classical gradient descent method needs $O(\epsilon^{-2})$ gradient evaluations, while GN-AGD($\cdot$) only needs $O(\epsilon^{-7/4} \log(1/\epsilon))$ gradient and function evaluations.

In each iteration, Algorithm 2 either performs GN-AGD($\cdot$) to find an $\epsilon$-first-order stationary point $\mathbf{x}_k$, i.e., $\|\nabla f(\mathbf{x}_k)\|_2 \leq \epsilon$, or takes a negative curvature step to escape from the small gradient region if $\mathbf{x}_{k-1}$ is a strict saddle point. In the extreme case that the objective function is convex or the algorithm does not encounter any strict saddle point with $\lambda_{\min}(\nabla^2 f(\mathbf{x})) < -\epsilon_H$, Algorithm 2 will perform negative curvature direction computation only once. This makes our algorithm very practical and appealing, compared with existing algorithms that heavily involve negative curvature computation in each iteration.

### 4.3 Runtime Analysis

Now we are going to present the runtime analysis of Algorithm 2. We first show that Algorithm 1 will escape from each saddle point in one step and satisfy the two conditions as we described in Section 4.1, i.e., function value decreases and gradient norm increases. It plays a pivotal role in the proof of the main theory.

**Lemma 4.3.** Suppose function $f(\cdot)$ is $L$-gradient Lipschitz and $\rho$-Hessian Lipschitz, $c_1 \geq 1$ and $\epsilon < \epsilon_H^2/(16c_1\rho)$, set $\eta = \epsilon_H/(2c_1\rho)$. If the input $\mathbf{x}$ of Algorithm 1 that satisfies $\|\nabla f(\mathbf{x})\|_2 \leq \epsilon$ and $\lambda_{\min}(\nabla^2 f(\mathbf{x})) < -\epsilon_H$, then with probability at least $1 - \delta$, the output $\mathbf{y}$ generated by Algorithm 1

satisfies
$$f(\mathbf{y}) - f(\mathbf{x}) < -\frac{C'\epsilon_H^3}{\rho^2}, \text{ and } \|\nabla f(\mathbf{y})\|_2 > \epsilon,$$

where $C' > 0$ is a constant, and the total runtime is $\widetilde{O}\big(\big[\sqrt{L/\epsilon_H}\,\big]\mathbb{T}_g\big)$.

The following lemma, which is adapted from Carmon et al. (2017), characterizes the iteration complexity of the GN-AGD algorithm.

**Lemma 4.4** (Carmon et al. (2017))**.** Let $f(\cdot)$ be a function that is $L$-gradient Lipschitz and $\rho$-Hessian Lipschitz continuous. Consider $\epsilon \in \big(0, \min\{\Delta_f^{2/3}\rho^{1/3}, L^2/(64\rho)\}\big)$, at the $k$-th outer loop of Algorithm 1, GN-AGD($\cdot$) returns a first-order stationary point $\mathbf{x}_k$ such that $\|\nabla f(\mathbf{x}_k)\|_2 < \epsilon$ with the total number of gradient evaluations

$$\widetilde{O}\bigg(\frac{\big(f(\mathbf{x}_{k-1}) - f(\mathbf{x}_k)\big)L^{1/2}\rho^{1/4}}{\epsilon^{7/4}} + \frac{L^{1/2}\rho^{-1/4}}{\epsilon^{1/4}}\bigg). \tag{4.3}$$

Based on Lemmas 4.3 and 4.4, we are able to perform the runtime complexity of Algorithm 2 for finding approximate local minima.

**Theorem 4.5.** Suppose function $f(\cdot)$ is $L$-gradient Lipschitz and $\rho$-Hessian Lipschitz. Setting $\epsilon < \epsilon_H^2/(16c_1\rho)$ with $c_1 \geq 1$, with probability $1 - \delta$, Algorithm 2 returns an $(\epsilon, \epsilon_H)$-second-order stationary point with runtime

$$\widetilde{O}\Bigg(\bigg[\frac{\Delta_f L^{1/2}\rho^{1/4}}{\epsilon^{7/4}} + \bigg(\frac{L^{1/2}\rho^{-1/4}}{\epsilon^{1/4}} + \frac{L^{1/2}}{\epsilon_H^{1/2}}\bigg)\min\bigg\{\frac{\rho^2\Delta_f}{\epsilon_H^3}, N_\epsilon\bigg\}\bigg]\mathbb{T}_g\Bigg), \tag{4.4}$$

where $N_\epsilon$ is the number of times Algorithm 2 enters small gradient regions.

**Remark 4.6.** Note that if we assume the runtime of gradient evaluation is in the same order as the runtime of Hessian-vector product, the computational complexity of FastPCA can be slightly better than Neon2 in the deterministic setting, since its iteration complexity has a better dependence in dimension $d$ by a factor of $\log d$.

**Remark 4.7.** It is worth noting that $N_\epsilon$ can be substantially smaller than $O(1/\epsilon_H^3)$, which implies the second term in (4.4) can be very small. Furthermore, if there only exists a finite number of strict saddle points, such as the function $f: \mathbb{R}^d \to \mathbb{R}$ constructed in Du et al. (2017), Algorithm 2 is guaranteed to estimate the negative curvature direction no more than $(d+1)$ times.

## 5 Stochastic Nonconvex Optimization

In this section, we consider the nonconvex optimization problem in the stochastic setting

$$\min_{\mathbf{x} \in \mathbb{R}^d} f(\mathbf{x}) = \mathbb{E}_{\xi \sim \mathcal{D}}[F(\mathbf{x}; \xi)], \tag{5.1}$$

where $\xi$ is a random variable satisfying distribution $\mathcal{D}$, $F(\mathbf{x}; \xi)$ is a stochastic smooth function and can be nonconvex. In the stochastic setting, one cannot directly access the full gradient and Hessian information of $f(\mathbf{x})$. Instead, one can only get unbiased estimators of the gradient and Hessian of $f(\mathbf{x})$. Note that the stochastic setting is referred to as online setting in some recent work (Allen-Zhu, 2017; Allen-Zhu and Li, 2017b).

Our goal is to find a local minimum of the expected function $f(\mathbf{x})$ in (5.1). Similar to the idea in the deterministic setting, we first need to find an $\epsilon$-first-order stationary point $\mathbf{x}_k$, and then take a negative curvature descent step to escape from the small gradient region if $\lambda_{\min}(\nabla^2 f(\mathbf{x}_k)) < -\epsilon_H$. Since we cannot access the entire information of $f(\mathbf{x})$, we will instead use a mini-batch of stochastic gradients and stochastic Hessian-vector products in the above procedures. In detail, we will first show that we could apply stochastic gradient and stochastic Hessian-based methods to estimate the negative curvature direction, and the resulting algorithm will still be able to escape from a saddle point by taking a single negative curvature descent step, which largely reduces the negative curvature computation. Then we integrate our stochastic one-step escaping algorithm and a stochastic gradient descent based algorithm to find local minima more efficiently.

## 5.1 Algorithm

In order to perform a negative curvature descent step as in Algorithm 1, by which the algorithm can escape from the small gradient region in one step, we need to find a negative curvature direction $\widehat{\mathbf{v}}$ in the stochastic setting. Similar to the deterministic setting, there are mainly two types of methods to find the negative curvature direction by only using stochastic gradient or stochastic Hessian vector product.

We can use online Oja's algorithm (Allen-Zhu and Li, 2017a) to compute the negative curvature direction in the stochastic setting, as shown in the following lemma.

**Lemma 5.1** (Allen-Zhu and Li (2017a)). Let $f(\mathbf{x}) = \mathbb{E}_\xi[F(\mathbf{x};\xi)]$ and $F(\mathbf{x};\xi)$ be a function that is $L$-gradient Lipschitz and $\rho$-Hessian Lipschitz continuous. For any $\mathbf{x} \in \mathbb{R}^d$, denote $\mathbf{H}$ as $\mathbf{H} = \nabla^2 f(\mathbf{x})$. If $\lambda_{\min}(\mathbf{H}) < -\epsilon_H$, then with probability at least $1 - \delta$, online Oja's algorithm will return a unit vector $\widehat{\mathbf{v}}$ satisfying
$$\widehat{\mathbf{v}}^\top \mathbf{H} \widehat{\mathbf{v}} < -\frac{\epsilon_H}{2},$$
in at most $O\big(\log^2(d/\delta)\log(1/\delta)L^2/\epsilon_H^2\big)$ stochastic Hessian-vector product evaluations.

We can also use online Neon2 (Allen-Zhu and Li, 2017b) to compute the negative curvature direction.

**Lemma 5.2** (Allen-Zhu and Li (2017b)). Let $f(\mathbf{x}) = \mathbb{E}_\xi[F(\mathbf{x};\xi)]$ and $F(\mathbf{x};\xi)$ be a function that is $L$-gradient Lipschitz and $\rho$-Hessian Lipschitz continuous. For any $\mathbf{x} \in \mathbb{R}^d$, denote $\mathbf{H}$ as $\mathbf{H} = \nabla^2 f(\mathbf{x})$. With probability at least $1 - \delta$, Neon2$^{\text{online}}$ returns $\widehat{\mathbf{v}}$ satisfying one of the following conditions,

- $\widehat{\mathbf{v}} = \bot$, then $\lambda_{\min}(\mathbf{H}) \geq -\epsilon_H$.

- $\widehat{\mathbf{v}} \neq \bot$, then $\widehat{\mathbf{v}}^\top \mathbf{H} \widehat{\mathbf{v}} \leq -\epsilon_H/2$ with $\|\widehat{\mathbf{v}}\|_2 = 1$.

The total number of stochastic gradient evaluations is $O\big(\log^2(d/\delta)L^2/\epsilon_H^2\big)$.

Based on the above two lemmas, in the stochastic setting, there exists an algorithm that uses stochastic gradient or stochastic Hessian vector product, denote by ApproxNC-Stochastic($\cdot$), and returns a unit vector $\widehat{\mathbf{v}}$ such that $\widehat{\mathbf{v}}^\top \nabla^2 f(\mathbf{x}) \widehat{\mathbf{v}} \leq -\epsilon_H/2$ if $\lambda_{\min}(\nabla^2 f(\mathbf{x})) < -\epsilon_H$, otherwise it returns $\widehat{\mathbf{v}} = \perp$.

Next we present our algorithm in Algorithm 3, which is able to escape from a saddle point in one step in the stochastic setting. Here $\nabla f_\mathcal{S}(\mathbf{x}) = 1/|\mathcal{S}| \sum_{\xi_i \in \mathcal{S}} \nabla F(\mathbf{x}; \xi_i)$ is a minibach of stochastic gradients.

---
**Algorithm 3** One-Step-Stochastic ($f(\cdot)$, $\mathbf{x}$, $\epsilon_H$, $\rho$, $\delta$)
---
1: Set $\eta \leftarrow \epsilon_H/(2c_1\rho)$, $|\mathcal{S}| = \widetilde{O}(1/\epsilon_H^2)$
2: $\widehat{\mathbf{v}} \leftarrow$ ApproxNC-Stochastic $(f(\cdot), \mathbf{x}, \epsilon_H, \delta)$
3: **if** $\widehat{\mathbf{v}} \neq \perp$ **then**
4:    $\widehat{\mathbf{g}} = \nabla f_\mathcal{S}(\mathbf{x})$
5:    $\widetilde{\mathbf{v}} \leftarrow \text{sign}(-\widehat{\mathbf{g}}^\top \widehat{\mathbf{v}}) \widehat{\mathbf{v}}$
6:    $\mathbf{y} = \mathbf{x} + \eta \widetilde{\mathbf{v}}$
7:    **return** $\mathbf{y}$
8: **else**
9:    **return** $\perp$
---

As shown in Algorithm 3, there are two differences compared with Algorithm 1. First, we adopt ApproxNC-Stochastic($\cdot$) here to find the negative curvature direction. Second, due to the fact that the full gradient of the expected function $f(\mathbf{x})$ is not accessible, we utilize a subsampled gradient $\widehat{\mathbf{g}}$ to approximate its full gradient, and then compute a descent direction accordingly, i.e., $\widetilde{\mathbf{v}}$. Since $\widehat{\mathbf{g}}$ is an unbiased estimator of $\nabla f(\mathbf{x})$, the difference between $\widehat{\mathbf{g}}$ and $\nabla f(\mathbf{x})$ can be sufficiently small as long as the sample size $|\mathcal{S}|$ is large enough. As a result, the negative curvature descent step along $\widetilde{\mathbf{v}}$ will also make the function value decrease and escape from the small gradient region if $\lambda_{\min}(\nabla^2 f(\mathbf{x})) \leq -\epsilon_H$.

To find local minima in the stochastic setting, similar to the deterministic setting, we need to first find an $\epsilon$-first-order stationary point $\mathbf{x}_k$ and then apply Algorithm 3 to escape the small gradient region or return the current iterate $\mathbf{x}_k$. Here we adopt the Stochastically Controlled Stochastic Gradient (SCSG) method in Lei et al. (2017), which is a variant of variance reduction based method for nonconvex optimization problems. The SCSG algorithm first estimates the full gradient using a minibach of stochastic gradients with batch size $B$, and then performs variance reduced semi-stochastic gradient with mini-batch size $b$ for $T_k$ times, where $T_k$ is randomly drawn from a geometric distribution with parameter $B/(B+b)$. The reason why we apply SCSG is that its iteration complexity to find an $\epsilon$-first-order stationary point is $O(1/\epsilon^{10/3})$, which is faster than other first-order stochastic optimization methods such as stochastic gradient descent (SGD) for general nonconvex optimization problems in the stochastic setting. Note that SCSG has been used in existing local minima finding algorithms (Xu and Yang, 2017; Allen-Zhu and Li, 2017b). Nevertheless, our algorithm is different from these algorithms in that our algorithm can further save the gradient and negative curvature evaluations due to the small gradient region and large gradient region division.

Then we present Algorithm 4 by combining One-Step-Stochastic($\cdot$) and the SCSG algorithm. In each outer loop, GOSE-Stochastic($\cdot$) either executes One-Step-Stochastic($\cdot$) or performs one-epoch SCSG algorithm according to a subsampled gradient measurement $\nabla f_{\mathcal{S}_k}(\mathbf{x}_{k-1})$. Note that we set the threshold as $\epsilon/2$ to ensure that the full gradient $\nabla f(\mathbf{x})$ is small when performing negative curvature descent step. If the norm of estimated gradient is large, i.e., $\|\nabla f_{\mathcal{S}_k}(\mathbf{x}_{k-1})\|_2 > \epsilon/2$, which means that the full gradient of $f(\mathbf{x})$ at $\mathbf{x}_{k-1}$ is also large according to the concentration result of $\nabla f_{\mathcal{S}_k}(\mathbf{x}_{k-1})$ on $\nabla f(\mathbf{x}_{k-1})$. Then we can show that $\mathbf{x}_k$ is not an $\epsilon$-first-order stationary point, and run the one-epoch SCSG algorithm. Otherwise, One-Step-Stochastic$(\cdot)$ should be performed in order to find the negative curvature direction and escape from the saddle point if $\lambda_{\min}(\nabla^2 f(\mathbf{x}_k)) < -\epsilon_H$. Note that in each iteration, Algorithm 4 either performs one-epoch SCSG or a negative curvature descent step, which can save the gradient and negative curvature computation to a large extent.

**Algorithm 4** GOSE-Stochastic $(f(\cdot), \mathbf{x}_0, \epsilon, \epsilon_H, L, \rho, \delta, K)$
───────────────────────────────────────────
1: Set $B \leftarrow \widetilde{O}(\mathcal{H}^*/\epsilon^2)$, $b \leftarrow \widetilde{O}(\rho^6 \mathcal{H}^* \epsilon^4 / L^3 \epsilon_H^9)$, $\eta \leftarrow b^{2/3}/6LB^{2/3}$
2: **for** $k = 1, 2, ..., K$
3:    uniformly sample a batch $\mathcal{S}_k \sim \mathcal{D}$ with $|\mathcal{S}_k| = B$
4:    $\mathbf{g}_k \leftarrow \nabla f_{\mathcal{S}_k}(\mathbf{x}_{k-1})$
5:    **if** $\|\mathbf{g}_k\|_2 > \epsilon/2$
6:      generate $T_k \sim \text{Geom}(B/(B+b))$
7:      $\mathbf{y}_0^{(k)} \leftarrow \mathbf{x}_{k-1}$
8:      **for** $t = 1, 2, ..., T_k$
9:        randomly pick $\mathcal{I}_{t-1} \subset [n]$ with $|\mathcal{I}_{t-1}| = b$
10:       $\mathbf{y}_t^{(k)} \leftarrow \mathbf{y}_{t-1}^{(k)} - \eta \big( \nabla f_{\mathcal{I}_{t-1}}(\mathbf{y}_{t-1}^{(k)}) - \nabla f_{\mathcal{I}_{t-1}}(\mathbf{y}_0^{(k)}) + \mathbf{g}_k \big)$
11:      **end for**
12:      $\mathbf{x}_k \leftarrow \mathbf{y}_{T_k}^{(k)}$
13:    **else**
14:      $\mathbf{x}_k \leftarrow $ One-Step-Stochastic$(f(\cdot), \mathbf{x}_{k-1}, \epsilon_H, \rho, \delta)$
15:      **if** $\mathbf{x}_k = \bot$
16:        return $\mathbf{x}_{k-1}$
17: **end for**
───────────────────────────────────────────

### 5.2 Runtime Analysis

Now we are going to provide a runtime analysis of Algorithm 4. Before we present our main result, we first make some additional assumptions on the stochastic function $F(\mathbf{x}; \xi)$.

**Assumption 5.3.** For all $\mathbf{x} \in \mathbb{R}^d$, $\xi \sim \mathcal{D}$, the stochastic gradient $\nabla F(\mathbf{x}; \xi)$ is sub-Gaussian with parameter $\sigma$, i.e., $\|\nabla F(\mathbf{x}; \xi) - \nabla f(\mathbf{x})\|_{\psi_2} \leq \sigma$, where $\|\mathbf{x}\|_{\psi_2} = \sup_{p \geq 1} p^{-1/2} \mathbb{E}\|\mathbf{x}\|_p$.

By Assumption 5.3, we immediately have

$$\mathbb{E}\|\nabla F(\mathbf{x}; \xi) - \nabla f(\mathbf{x})\|_2^2 \leq 2\sigma^2 \triangleq \mathcal{H}^*,$$

which is a uniform upper bound of the variance of $F(\mathbf{x}; \xi)$ for all $\mathbf{x} \in \mathbb{R}^d$.

Next, we give a large deviation bound on the distance between the subsampled gradient $\widehat{\mathbf{g}}$ and the full gradient $\nabla f(\mathbf{x})$ in the following lemma.

**Lemma 5.4.** (Ghadimi et al., 2016) Suppose the stochastic gradient $\nabla F(\mathbf{x}; \xi)$ is sub-Gaussian, for any given $c > 0$, if the sample size $|\mathcal{S}| = O((\sigma^2/c^2\epsilon^2)\log(1/\delta))$, then with probability $1 - \delta$,

$$\|\widehat{\mathbf{g}} - \nabla f(\mathbf{x})\|_2 \leq c\,\epsilon$$

holds for any $\mathbf{x} \in \mathbb{R}^d$, where $\widehat{\mathbf{g}} = 1/|\mathcal{S}| \sum_{\xi_i \in \mathcal{S}} \nabla F(\mathbf{x}; \xi_i)$.

The above lemma is a standard concentration bound for sub-Gaussian random vectors (Vershynin, 2010; Ghadimi et al., 2016), which implies that we can use the subsampled gradient $\widehat{\mathbf{g}}$ to approximate the full gradient $\nabla f(\mathbf{x})$ provided that the subsample size is large enough.

Next we show that Algorithm 3 can escape from each saddle point in one step and satisfy the two conditions as we described in Section 4.1, i.e., function value decreases and gradient norm increases.

**Lemma 5.5.** Let $f(\mathbf{x}) = \mathbb{E}_\xi[F(\mathbf{x}; \xi)]$ and suppose $F(\mathbf{x}; \xi)$ be a function that is $L$-smooth and $\rho$-Hessian Lipschitz continuous, $c_1 \geq 1$ and $\epsilon < \epsilon_H^2/(16c_1\rho)$, $|\mathcal{S}| = \widetilde{O}(1/\epsilon_H^2)$, and set step size $\eta = \epsilon_H/(2c_1\rho)$. If the input $\mathbf{x}$ of Algorithm 3 that statisfies $\|\nabla f(\mathbf{x})\|_2 \leq \epsilon$ and $\lambda_{\min}(\nabla^2 f(\mathbf{x})) < -\epsilon_H$, then with probability at least $1 - \delta$, the output $\mathbf{y}$ generated by Algorithm 3 satisfies

$$f(\mathbf{y}) - f(\mathbf{x}) < -\frac{C'\epsilon_H^3}{\rho^2}, \text{ and } \|\nabla f(\mathbf{y})\|_2 > \epsilon,$$

where $C' > 0$ is a constant, and the runtime is $\widetilde{O}\big([L^2/\epsilon_H^2]\mathbb{T}_g\big)$.

The above lemma is similar to Lemma 4.3 in the deterministic setting. Note that here we need to deal with the subsampled gradient rather than the exact one, and the runtime is different from that in Lemma 4.3.

In what follows, a crucial lemma that characterizes the function value increments for one-epoch SCSG algorithm is presented.

**Lemma 5.6** (Lei et al. (2017)). Consider $f(\mathbf{x}) = \mathbb{E}_\xi[F(\mathbf{x}; \xi)]$, assume that $F(\mathbf{x}; \xi)$ is $L$-smooth Lipschitz continuous, and the stochastic gradient $\nabla F(\mathbf{x}; \xi)$ is sub-Gaussian. Suppose that Algorithm 4 performs one-epoch SCSG algorithm at $k$-th outer loop, there exists constant $C > 1$ such that the following holds for $\mathbf{y}_0^{(k)}$ and $\mathbf{y}_{T_k}^{(k)}$,

$$\mathbb{E}\big[\|\nabla f(\mathbf{y}_{T_k}^{(k)})\|_2^2\big] \leq \frac{CLb^{1/3}}{B^{1/3}}\mathbb{E}\big[f(\mathbf{y}_0^{(k)}) - f(\mathbf{y}_{T_k}^{(k)})\big] + \frac{6 \cdot \mathbb{1}\{B < n\}}{B} \cdot \mathcal{H}^*,$$

where $B$ and $b$ denote the batch size and mini-batch size respectively, and $\mathcal{H}^*$ is an upper bound on the variance of stochastic gradients.

Based on the above two lemmas, we are able to establish the runtime complexity guarantee of Algorithm 4 with constant probability.

**Theorem 5.7.** Consider $f(\mathbf{x}) = \mathbb{E}_\xi[F(\mathbf{x}; \xi)]$, assume that $F(\mathbf{x}; \xi)$ is $L$-smooth and $\rho$-Hessian Lipschitz continuous, the stochastic gradient $\nabla F(\mathbf{x}; \xi)$ is sub-Gaussian, and $\epsilon \leq \min\big\{\epsilon_H^{3/2}, \epsilon_H^2/(16c_1\rho)\big\}$. Considering batch size $B = \widetilde{O}(\mathcal{H}^*/\epsilon^2)$, mini-batch size $b = \widetilde{O}(\rho^6\mathcal{H}^*\epsilon^4/(L^3\epsilon_H^9))$, and $\eta = b^{2/3}/(6LB^{2/3})$, with probability $1/3$, Algorithm 4 returns an $(\epsilon,\epsilon_H)$-second-order stationary point with runtime

$$\widetilde{O}\bigg(\bigg[\frac{L\Delta_f\mathcal{H}^{*2/3}}{\epsilon^{10/3}} + \frac{\rho^2\Delta_f\mathcal{H}^*}{\epsilon_H^3\epsilon^2} + \bigg(\frac{\mathcal{H}^*}{\epsilon^2} + \frac{L^2}{\epsilon_H^2}\bigg)\min\bigg\{\frac{\rho^2\Delta_f}{\epsilon_H^3}, N_\epsilon\bigg\}\bigg]\mathbb{T}_g\bigg),$$

where $N_\epsilon$ is the number of times Algorithm 4 enters small gradient regions.

**Remark 5.8.** Note that the runtime complexity guarantee in Theorem 5.7 holds with probability $1/3$. In practice, one can improve the constant probability to $1 - \delta$ probability for any $0 < \delta < 1$ by repeat GOSE-Stochastic for $O(\log(1/\delta))$ times.

**Remark 5.9.** We consider a special regime that $\epsilon_H \lesssim \epsilon^{2/3}$, which implies that $b \geq B$. Then SCSG will degenerate into the SGD algorithm. Since the derivation for its runtime complexity is similar to that for SCSG, we omit the analysis for SGD in this work. The interested readers can refer to Xu and Yang (2017); Allen-Zhu and Li (2017b) for more details. The rest of the discussion is in the regime that $\epsilon_H \gtrsim \epsilon^{2/3}$, i.e., $b < B$. In this regime, if the number of encountered saddle points in Algorithm 4 is smaller than $O(\epsilon_H^{-3})$, GOSE-Stochastic($\cdot$) outperforms the state-of-the-art algorithms (Tripuraneni et al., 2017; Xu and Yang, 2017; Allen-Zhu and Li, 2017b). In addition, when $\mathcal{H}^* = 0$, SCSG degenerates to full gradient descent method, and the number of stochastic gradient evaluations for finding an $(\epsilon, \epsilon_H)$-second-order stationary point becomes $\widetilde{O}(L\Delta_f \epsilon^{-2} + L^2 \epsilon_H^{-2} \min\{\rho^2 \Delta_f \epsilon_H^{-3}, N_\epsilon\})$.

# 6 Finite-Sum Nonconvex Optimization

In this section, we consider a special case of nonconvex stochastic optimization, where the objective function can be written as an average of finite sum component functions

$$\min_{\mathbf{x} \in \mathbb{R}^d} f(\mathbf{x}) = \frac{1}{n} \sum_{i=1}^n f_i(\mathbf{x}), \tag{6.1}$$

where each $f_i(\mathbf{x})$ is smooth and can be nonconvex. Nonconvex optimization problems with this generic structure appear widely in machine learning such as training deep neural networks (LeCun et al., 2015). In contrast to the stochastic setting we discussed before, here we have access to the full information of function $f(\cdot)$, which is referred to the offline setting in some exiting work (Allen-Zhu, 2017; Tripuraneni et al., 2017).

Compared with the deterministic setting, it has been shown that stochastic algorithms can utilize the finite-sum structure in (6.1) and further reduce the gradient complexity for both convex and nonconvex optimization problems. As for nonconvex finite-sum optimization problems, previous work (Reddi et al., 2016; Allen-Zhu and Hazan, 2016) showed that nonconvex stochastic variance reduced gradient (SVRG) methods are faster than classical gradient descent methods by a factor of $O(n^{1/3})$. As a result, we could make use of the finite-sum structure and further reduce the runtime complexity of Algorithm 2 if the objective function is of finite sum structure.

## 6.1 Algorithm

In order to escape from each saddle point in one step to reduce the negative curvature computation, first we need to find a negative curvature direction $\widehat{\mathbf{v}}$ in the finite-sum setting. Similar to the previous settings, there exist two stochastic methods to find the negative curvature direction by using gradient or Hessian-vector product evaluations. We present the runtime complexity results of these two methods in the following two lemmas respectively.

**Lemma 6.1** (Garber et al. (2016)). Let $f(\mathbf{x}) = \frac{1}{n}\sum_{i=1}^{n} f_i(\mathbf{x})$ and each $f_i(\cdot)$ be a function that is $L$-smooth and $\rho$-Hessian Lipschitz continuous. For any $\mathbf{x} \in \mathbb{R}^d$, denote $\mathbf{H}$ as $\mathbf{H} = \nabla^2 f(\mathbf{x})$. Then with probability at least $1 - \delta$, Shifted-and-Inverted power method will return a unit vector $\widehat{\mathbf{v}}$ satisfying

$$\widehat{\mathbf{v}}^\top \mathbf{H} \widehat{\mathbf{v}} < \lambda_{\min}(\mathbf{H}) + \frac{\epsilon_H}{2},$$

in at most $O\big((n + n^{3/4}\sqrt{L/\epsilon_H})\log^3(d)\log(1/\delta)\big)$ stochastic Hessian-vector product evaluations.

**Lemma 6.2** (Allen-Zhu and Li (2017b)). Let $f(\mathbf{x}) = \frac{1}{n}\sum_{i=1}^{n} f_i(\mathbf{x})$ and each $f_i(\cdot)$ be a function that is $L$-smooth and $\rho$-Hessian Lipschitz continuous. For any $\mathbf{x} \in \mathbb{R}^d$, denote $\mathbf{H}$ as $\mathbf{H} = \nabla^2 f(\mathbf{x})$. With probability at least $1 - \delta$, Neon2$^{\text{svrg}}$ returns $\widehat{\mathbf{v}}$ satisfying one of the following conditions,

- $\widehat{\mathbf{v}} = \bot$, then $\lambda_{\min}(\mathbf{H}) \geq -\epsilon_H$.
- $\widehat{\mathbf{v}} \neq \bot$, then $\widehat{\mathbf{v}}^\top \mathbf{H} \widehat{\mathbf{v}} \leq -\epsilon_H/2$ with $\|\mathbf{v}\|_2 = 1$.

The total number of stochastic gradient evaluations is $O\big((n + n^{3/4}\sqrt{L/\epsilon_H})\log^2(d/\delta)\big)$.

Different from the deterministic setting, the runtime complexities of these two methods for finding the negative curvature direction are faster than those in the deterministic setting by a factor of $O(n^{1/4})$, because they can exploit the finite-sum structure. According to above two lemmas, in the finite-sum setting, there exists an algorithm, denoted by ApproxNC-FiniteSum($\cdot$), that only uses either gradient evaluation, or Hessian vector product, and returns a unit vector $\widehat{\mathbf{v}}$ such that $\widehat{\mathbf{v}}^\top \nabla^2 f(\mathbf{x}) \widehat{\mathbf{v}} \leq -\epsilon_H/2$ if $\lambda_{\min}(\nabla^2 f(\mathbf{x})) < -\epsilon_H$, otherwise it returns $\widehat{\mathbf{v}} = \bot$. Based on ApproxNC-FiniteSum($\cdot$), we present Algorithm 5, which is able to escape from each saddle point in one step in the finite-sum setting.

---

**Algorithm 5** One-Step-FiniteSum $(f(\cdot), \mathbf{x}, \epsilon_H, \rho, \delta)$

1: Set $\eta \leftarrow \epsilon_H/2c_1\rho$
2: $\widehat{\mathbf{v}} \leftarrow$ ApproxNC-FiniteSum $(f(\cdot), \mathbf{x}, \epsilon_H, \delta)$
3: **if** $\widehat{\mathbf{v}} \neq \bot$ **then**
4: $\quad \widetilde{\mathbf{v}} \leftarrow \text{sign}(-\nabla f(\mathbf{x})^\top \widehat{\mathbf{v}})\widehat{\mathbf{v}}$
5: $\quad \mathbf{y} = \mathbf{x} + \eta \widetilde{\mathbf{v}}$
6: $\quad$ **return** $\mathbf{y}$
7: **else**
8: $\quad$ **return** $\bot$

---

The remaining thing is similar to previous settings. In detail, to find $\epsilon$-first-order stationary points, we adopt the nonconvex stochastic variance reduced gradient (SVRG) method (Reddi et al., 2016; Allen-Zhu and Hazan, 2016; Lei et al., 2017) in our algorithm, since it is $O(n^{1/3})$ faster than classical gradient descent methods in the nonconvex finite-sum setting. For the simplicity of presentation, we adopt the Stochastically Controlled Stochastic Gradient (SCSG) method (Lei et al., 2017), a variant of SVRG, as an instance of our algorithm, while other algorithms in this family are also applicable and will give rise to the same theoretical guarantee. In order to escape from the small gradient region in one step, we call Algorithm 5.

**Algorithm 6** GOSE-FiniteSum ($f(\cdot)$, $\mathbf{x}_0$, $\epsilon$, $\epsilon_H$, $L$, $\rho$, $\delta$, $K$)

1: Set $b \leftarrow 1$, $\eta \leftarrow 1/Ln^{2/3}$
2: **for** $k = 1, 2, ..., K$
3:     $\mathbf{g}_k \leftarrow \nabla f(\mathbf{x}_{k-1})$
4:     **if** $\|\mathbf{g}_k\|_2 > \epsilon$
5:         generate $T_k \sim \text{Geom}(n/(n+b))$
6:         $\mathbf{y}_0^{(k)} \leftarrow \mathbf{x}_{k-1}$
7:         **for** $t = 1, 2, ..., T_k$
8:             randomly pick $\mathcal{I}_{t-1} \subset [n]$ with $|\mathcal{I}_{t-1}| = b$
9:             $\mathbf{y}_t^{(k)} \leftarrow \mathbf{y}_{t-1}^{(k)} - \eta\big(\nabla f_{\mathcal{I}_{t-1}}(\mathbf{y}_{t-1}^{(k)}) - \nabla f_{\mathcal{I}_{t-1}}(\mathbf{y}_0^{(k)}) + \mathbf{g}_k\big)$
10:        **end for**
11:        $\mathbf{x}_k \leftarrow \mathbf{y}_{T_k}^{(k)}$
12:     **else**
13:        $\mathbf{x}_k \leftarrow$ One-Step-FiniteSum($f(\cdot), \mathbf{x}_{k-1}, \epsilon_H, \rho, \delta$)
14:        **if** $\mathbf{x}_k = \bot$
15:            **return** $\mathbf{x}_{k-1}$
16: **end for**

We present GOSE-FiniteSum($\cdot$) in Algorithm 6, which either performs One-epoch SCSG algorithm or One-Step-FiniteSum algorithm, depending on the norm of gradient $\nabla f(\mathbf{x}_{k-1})$ in the beginning of each outer loop. Compared with Algorithm 4 in the stochastic setting, here $\nabla f(\mathbf{x}_{k-1})$ is the full gradient at $\mathbf{x}_{k-1}$. Algorithm 6 keeps performing One-epoch SCSG until it finds an $\epsilon$-first-order stationary point. Then it will take a negative curvature descent step if $\lambda_{\min}(\nabla^2 f(\mathbf{x}_{k-1})) < -\epsilon_H$, otherwise it will return $\mathbf{x}_{k-1}$. Similar to the algorithms in deterministic and stochastic settings, Algorithm 6 reduces gradient and negative curvature computation as much as possible, and can be very efficient in practice.

## 6.2 Runtime Analysis

Now we provide the theoretical analysis of Algorithms 5 and 6. To begin with, we present the runtime analysis of Algorithm 5.

**Lemma 6.3.** Suppose function $f(\cdot)$ is $L$-gradient Lipschitz and $\rho$-Hessian Lipschitz, $c_1 \geq 1$ and $\epsilon < \epsilon_H^2/(16c_1\rho)$, set step size $\eta = \epsilon_H/(2c_1\rho)$. If the $\mathbf{x}$ of Algorithm 5 that satisfies $\|\nabla f(\mathbf{x})\|_2 \leq \epsilon$ and $\lambda_{\min}(\nabla^2 f(\mathbf{x})) < -\epsilon_H$, then with probability at least $1-\delta$, the output $\mathbf{y}$ generated by Algorithm 1 satisfies

$$f(\mathbf{y}) - f(\mathbf{x}) < -\frac{C'\epsilon_H^3}{\rho^2}, \text{ and } \|\nabla f(\mathbf{y})\|_2 > \epsilon,$$

where $C' > 0$ is a constant, and the runtime is $\widetilde{O}\big([n + n^{3/4}\sqrt{L/\epsilon_H}\,]\mathbb{T}_g\big)$.

The proof of the above lemma is almost identical to Lemma 4.3, except that the runtime complexity of Algorithm 5 is improved over Algorithm 1. To analyze Algorithm 6, we first present the following lemma, which characterizes the expected function value increments when performing one-epoch SCSG algorithm.

**Lemma 6.4** (Lei et al. (2017)). Suppose function $f(\cdot)$ is $L$-gradient Lipschitz, suppose Algorithm 6 perform one-epoch SCSG algorithm at $k$-th outer loop, there exists constant $C > 1$ such that the following holds for $\mathbf{y}_0^{(k)}$ and $\mathbf{y}_{T_k}^{(k)}$

$$\mathbb{E}\big[\|\nabla f(\mathbf{y}_{T_k}^{(k)})\|_2^2\big] \leq CL(b/B)^{1/3}\mathbb{E}\big[f(\mathbf{y}_0^{(k)}) - f(\mathbf{y}_{T_k}^{(k)})\big],$$

where $B$ and $b$ denote the batch size and mini-batch size, respectively.

It is worth noting that similar results as Lemma 6.4 have also been derived in Reddi et al. (2016); Allen-Zhu and Hazan (2016). Now, based on Lemma 6.3 and Lemma 6.4, we are ready to deliver the main result for Algorithm 6 as follows.

**Theorem 6.5.** Suppose function $f(\cdot)$ is $L$-gradient Lipschitz and $\rho$-Hessian Lipschitz. Considering the batch size $B = n$, mini-batch size $b = 1$, and $\epsilon \leq \epsilon_H^2/(16c_1\rho)$, with probability $1/3$, Algorithm 6 returns an $(\epsilon, \epsilon_H)$-second-order stationary point with runtime

$$\widetilde{O}\bigg(\bigg[\frac{L\Delta_f n^{2/3}}{\epsilon^2} + \bigg(n + \frac{n^{3/4}L^{1/2}}{\epsilon_H^{1/2}}\bigg)\min\bigg\{\frac{\rho^2\Delta_f}{\epsilon_H^3}, N_\epsilon\bigg\}\bigg]\mathbb{T}_g\bigg),$$

where $N_\epsilon$ is the number of times Algorithm 6 enters small gradient regions.

**Remark 6.6.** Note that the runtime complexity in Theorem 6.5 holds with probability $1/3$. One can improve it to probability $1 - \delta$ by randomly repeat GOSE-Stochastic for $O(\log(1/\delta))$ times. In addition, when the total sample size $n \gtrsim \epsilon^{-3/2}$ and $N_\epsilon$ is substantially less than $O(\epsilon_H^{-3})$, the total runtime of Algorithm 6 is $\widetilde{O}([n^{2/3}\epsilon^{-2}]\mathbb{T}_g)$, which outperforms the best finite-sum algorithms in Agarwal et al. (2016); Allen-Zhu and Li (2017b).

# 7 Conclusions and Future Work

We proposed a family of algorithms for finding approximate local minima for general nonconvex optimization in different settings. Our algorithms are guaranteed to escape from each saddle point in one step using negative curvature descent, and save gradient and negative curvature computation to a large extent. An interesting question is whether our algorithm can be extended to constrained nonconvex optimization problems. We will study it in the future work.

# A Proofs for Deterministic Nonconvex Optimization

## A.1 Proof of Lemma 4.3

*Proof.* First, we show that the negative curvature step (4.2) will decrease the function value, i.e., $f(\mathbf{y}) - f(\mathbf{x}) < 0$. Based on the assumption that $f(\cdot)$ is $\rho$-Hessian Lipschitz, we can get

$$f(\mathbf{y}) \leq f(\mathbf{x}) + \langle\nabla f(\mathbf{x}), \mathbf{y} - \mathbf{x}\rangle + \frac{1}{2}\langle\mathbf{y} - \mathbf{x}, \nabla^2 f(\mathbf{x})(\mathbf{y} - \mathbf{x})\rangle + \frac{\rho}{6}\|\mathbf{y} - \mathbf{x}\|_2^3, \ \forall \mathbf{x}, \mathbf{y} \in \mathbb{R}^d.$$

According to the update form $\mathbf{y} = \mathbf{x} + \eta \widetilde{\mathbf{v}}$, we could get

$$f(\mathbf{y}) \leq f(\mathbf{x}) + \eta \langle \nabla f(\mathbf{x}), \widetilde{\mathbf{v}} \rangle + \frac{\eta^2}{2} \langle \widetilde{\mathbf{v}}, \nabla^2 f(\mathbf{x}) \widetilde{\mathbf{v}} \rangle + \frac{\rho \eta^3}{6} \|\widetilde{\mathbf{v}}\|_2^3.$$

Based on the definition of $\mathbf{v}_t$, we obtain

$$\langle \nabla f(\mathbf{x}), \widetilde{\mathbf{v}} \rangle = \text{sign}(-\nabla f(\mathbf{x})^\top \widehat{\mathbf{v}}) \langle \nabla f(\mathbf{x}), \widehat{\mathbf{v}} \rangle \leq 0.$$

Then we have

$$f(\mathbf{y}) \leq f(\mathbf{x}) + \frac{\eta^2}{2} \langle \widetilde{\mathbf{v}}, \nabla^2 f(\mathbf{x}) \widetilde{\mathbf{v}} \rangle + \frac{\rho \eta^3}{6} \|\widetilde{\mathbf{v}}\|_2^3$$
$$\leq f(\mathbf{x}) - \left( \frac{\epsilon_H \eta^2}{4} - \frac{\rho \eta^3}{6} \right).$$

Taking $\eta = C_H (\epsilon_H/\rho)$ such that

$$0 < \eta < \frac{3\epsilon_H}{2\rho}, \tag{A.1}$$

which implies that $C_H \in (0, 3/2)$, we are able to obtain

$$f(\mathbf{y}) - f(\mathbf{x}) \leq -\left( \frac{C_H^2}{4} - \frac{C_H^3}{6} \right) \frac{\epsilon_H^3}{\rho^2} < 0,$$

which guarantees that there will be a sufficient decrease after performing negative curvature step (4.2).

Secondly, we prove that the norm of the gradient increases after the negative curvature step, i.e., $\|\nabla f(\mathbf{y})\|_2 > \epsilon$. Note that the gradient $\nabla f(\mathbf{x}_{k+1})$ can be rewritten as:

$$\nabla f(\mathbf{y}) = \nabla f(\mathbf{x}) + \eta \int_0^1 \nabla^2 f(\mathbf{x} + \theta \eta \widetilde{\mathbf{v}}) \widetilde{\mathbf{v}} d\theta$$
$$= \nabla f(\mathbf{x}) + \eta \int_0^1 \nabla^2 f(\mathbf{x}) \widetilde{\mathbf{v}} d\theta + \eta' \int_0^1 \left( \nabla^2 f(\mathbf{x} + \theta \eta \widetilde{\mathbf{v}}) - \nabla^2 f(\mathbf{x}) \right) \widetilde{\mathbf{v}} d\theta$$
$$= \nabla f(\mathbf{x}_t) + \eta' \nabla^2 f(\mathbf{x}_t) \mathbf{v}_t + \mathbf{\Delta}_t,$$

where $\mathbf{\Delta}_t = \eta \int_0^1 \left( \nabla^2 f(\mathbf{x} + \theta \eta \widetilde{\mathbf{v}}) - \nabla^2 f(\mathbf{x}) \right) \widetilde{\mathbf{v}} d\theta$. Then we analyze the upper bound of $\|\mathbf{\Delta}_t\|_2$,

$$\|\mathbf{\Delta}_t\|_2 = \eta \left\| \int_0^1 \left( \nabla^2 f(\mathbf{x} + \theta \eta \widetilde{\mathbf{v}}) - \nabla^2 f(\mathbf{x}) \right) \widetilde{\mathbf{v}} d\theta \right\|_2$$
$$\leq \eta \int_0^1 \left\| \nabla^2 f(\mathbf{x} + \theta \eta \widetilde{\mathbf{v}}) - \nabla^2 f(\mathbf{x}) \right\|_2 \|\widetilde{\mathbf{v}}\|_2 d\theta$$
$$\leq \rho \eta^2 \int_0^1 \theta \|\widetilde{\mathbf{v}}\|_2^2 d\theta$$
$$= \frac{1}{2} \rho \eta^2.$$

Next we get the lower bound of $\|\nabla f(\mathbf{y}) - \nabla f(\mathbf{x})\|_2$,

$$\|\nabla f(\mathbf{y}) - \nabla f(\mathbf{x})\|_2 \geq \eta\|\nabla^2 f(\mathbf{x})\widetilde{\mathbf{v}}\|_2 - \frac{1}{2}\rho\eta^2$$

$$= \eta\|\widetilde{\mathbf{v}}\|_2 \cdot \|\nabla^2 f(\mathbf{x})\widetilde{\mathbf{v}}\|_2 - \frac{1}{2}\rho\eta^2$$

$$\geq \eta|\langle\widetilde{\mathbf{v}}, \nabla^2 f(\mathbf{x})\widetilde{\mathbf{v}}\rangle| - \frac{1}{2}\rho\eta^2$$

$$\geq \frac{1}{2}\epsilon_H\eta - \frac{1}{2}\rho\eta^2,$$

where the first equation is because $\|\widetilde{\mathbf{v}}\|_2 = 1$, and the third inequality is because $|\widetilde{\mathbf{v}}^\top \nabla^2 f(\mathbf{x})\widetilde{\mathbf{v}}| \geq \epsilon_H/2$. Then we can derive that

$$\|\nabla f(\mathbf{y})\|_2 \geq \frac{1}{2}\epsilon_H\eta - \frac{1}{2}\rho\eta^2 - \|\nabla f(\mathbf{x})\|_2 \geq \frac{1}{2}\epsilon_H\eta - \frac{1}{2}\rho\eta^2 - \epsilon.$$

In order to guarantee that $\|f(\mathbf{y})\|_2 > \epsilon$, we need to set $\eta$ satisfies

$$\frac{1}{2}\epsilon_H\eta - \frac{1}{2}\rho\eta^2 - \epsilon > \epsilon \Leftrightarrow \frac{1}{2}\epsilon_H\eta - \frac{1}{2}\rho\eta^2 - 2\epsilon > 0.$$

Moreover, it is necessary to set $\epsilon < \epsilon_H^2/16\rho$ in order to guarantee there exists a solution to the above inequality, then we can get

$$\frac{\epsilon_H}{2\rho} - \sqrt{\frac{\epsilon_H^2}{4\rho^2} - \frac{4\epsilon}{\rho}} < \eta < \frac{\epsilon_H}{2\rho} + \sqrt{\frac{\epsilon_H^2}{4\rho^2} - \frac{4\epsilon}{\rho}}.$$

As we defined $\eta$ as $\eta = C_H(\epsilon_H/\rho)$, the above inequality is equivalent to

$$\frac{1}{2} - \frac{1}{2}\sqrt{1 - \frac{16\rho\epsilon}{\epsilon_H^2}} < C_H < \frac{1}{2} + \frac{1}{2}\sqrt{1 - \frac{16\rho\epsilon}{\epsilon_H^2}}. \qquad (\text{A.2})$$

It is worth noting that if $C_H$ satisfies the above condition (A.1), it also satisfies condition (A.2). Finally we can obtain that if $\eta = C_H(\epsilon_H/\rho)$, and $C_H \in (1/2 - 1/2\sqrt{1 - 16\rho\epsilon/\epsilon_H^2}, 1/2 + 1/2\sqrt{1 - 16\rho\epsilon/\epsilon_H^2})$, the negative curvature step (4.2) will satisfies

$$f(\mathbf{y}) - f(\mathbf{x}) < -\frac{C'\epsilon_H^3}{\rho^2}, \text{ and } \|\nabla f(\mathbf{y})\|_2 > \epsilon,$$

where $C' = C_H^2/4 - C_H^3/6 > 0$ is a constant. For choosing the step size parameter $\eta$, we can overshoot the Hessian Lipschitz parameter $\rho$ as $c_1\rho$ where $c_1 \geq 1$. Then the conclusion still holds and we can simply choose the step size parameter as $\eta = \epsilon_H/2c_1\rho$, where we take $C_H = 1/2$ and replace $\rho$ with $c_1\rho$, and this concludes our proof. $\square$

## A.2 Proof of Theorem 4.5

*Proof.* By Lemma 4.4, we know that in the line 4 of Algorithm 2, GN-AGD($\cdot$) outputs $\mathbf{x}_{k+1}$ with the number of gradient evaluations

$$T_k^G = \widetilde{O}\bigg(\frac{\big(f(\mathbf{x}_k) - f(\mathbf{x}_{k+1})\big)L^{1/2}\rho^{1/4}}{\epsilon^{7/4}} + \frac{L^{1/2}\rho^{-1/4}}{\epsilon^{1/4}}\bigg)$$

In addition, by Lemmas 4.1 and 4.2, it can be seen that the complexity of computing approximate negative curvature is in the order of $\widetilde{O}(\sqrt{L/\epsilon_H})$, and the number of calls to One-Step-Deterministic function is upper bounded by $\widetilde{O}\big(\min\{\rho^2\Delta_f/\epsilon_H^3, N_\epsilon\}\big)$. Thus, let $\mathcal{G}$ denote the iteration set that the outer loop that GN-AGD($\cdot$) is executed, and $\mathcal{G}^c$ denote the iteration set that performing One-Step-Deterministic($\cdot$), the total runtime complexity of GOSE-Deterministic can be obtained by performing the summation over the complexity in each outer loop, i.e.,

$$T = \widetilde{O}\bigg(\bigg[\frac{\sum_{k\in\mathcal{G}}\big[f(\mathbf{x}_k) - f(\mathbf{x}_{k+1})\big]L^{1/2}\rho^{1/4}}{\epsilon^{7/4}} + \sum_{k\in\mathcal{G}^c}\bigg[\frac{L^{1/2}\rho^{-1/4}}{\epsilon^{1/4}} + \frac{L^{1/2}}{\epsilon_H^{1/2}}\bigg]\bigg]\mathbb{T}_g\bigg)$$

$$= \widetilde{O}\bigg(\bigg[\frac{\sum_{k\in\mathcal{G}}\big[f(\mathbf{x}_k) - f(\mathbf{x}_{k+1})\big]L^{1/2}\rho^{1/4}}{\epsilon^{7/4}} + \bigg(\frac{L^{1/2}\rho^{-1/4}}{\epsilon^{1/4}} + \frac{L^{1/2}}{\epsilon_H^{1/2}}\bigg)\min\bigg\{\frac{\rho^2\Delta_f}{\epsilon_H^3}, N_\epsilon\bigg\}\bigg]\mathbb{T}_g\bigg)$$

$$= \widetilde{O}\bigg(\bigg[\frac{\Delta_f L^{1/2}\rho^{1/4}}{\epsilon^{7/4}} + \bigg(\frac{L^{1/2}\rho^{-1/4}}{\epsilon^{1/4}} + \frac{L^{1/2}}{\epsilon_H^{1/2}}\bigg)\min\bigg\{\frac{\rho^2\Delta_f}{\epsilon_H^3}, N_\epsilon\bigg\}\bigg]\mathbb{T}_g\bigg),$$

where the third equality follows from the fact $\sum_{k\in\mathcal{G}} f(\mathbf{x}_k) - f(\mathbf{x}_{k+1}) \leq \Delta_f$, since One-Step-Deterministic($\cdot$) is able to guarantee the non-increasing property in terms of the function value. $\square$

## B Proofs for Stochastic Nonconvex Optimization

### B.1 Proof of Lemma 5.5

*Proof.* To begin with, we show that the negative curvature descent step will decrease the function value. Based on the assumption that $f(\cdot)$ is $\rho$-Hessian Lipschitz, we can get

$$f(\mathbf{y}) \leq f(\mathbf{x}) + \langle\nabla f(\mathbf{x}), \mathbf{y} - \mathbf{x}\rangle + \frac{1}{2}\langle\mathbf{y} - \mathbf{x}, \nabla^2 f(\mathbf{x})(\mathbf{y} - \mathbf{x})\rangle + \frac{\rho}{6}\|\mathbf{y} - \mathbf{x}\|_2^3, \ \forall\, \mathbf{x}, \mathbf{y} \in \mathbb{R}^d.$$

Since $\lambda_{\min}(\nabla^2 f(\mathbf{x}_t)) < -\epsilon_H$, we could get the negative curvature direction $\widehat{\mathbf{v}}$ such that

$$\widehat{\mathbf{v}}^\top \nabla^2 f(\mathbf{x})\widehat{\mathbf{v}} \leq -\frac{\epsilon_H}{2}, \quad \|\widehat{\mathbf{v}}\|_2 = 1,$$

which leads to the update

$$\mathbf{y} = \mathbf{x} + \eta \cdot \text{sign}(-\widehat{\mathbf{g}}^\top\widehat{\mathbf{v}})\widehat{\mathbf{v}} = \mathbf{x}_t + \eta\,\widetilde{\mathbf{v}},$$

where $\widehat{\mathbf{g}} = 1/|S|\sum_{\xi_i \in S} \nabla F(\mathbf{x}; \xi_i)$. Based on the above inequalities and update form, we have

$$f(\mathbf{y}) \leq f(\mathbf{x}) + \eta\langle\nabla f(\mathbf{x}), \widetilde{\mathbf{v}}\rangle + \frac{\eta^2}{2}\langle\widetilde{\mathbf{v}}, \nabla^2 f(\mathbf{x})\widetilde{\mathbf{v}}\rangle + \frac{\rho\eta^3}{6}\|\widetilde{\mathbf{v}}\|_2^3.$$

According to the definition of $\widetilde{\mathbf{v}}$, we obtain

$$\langle \widehat{\mathbf{g}}, \widetilde{\mathbf{v}} \rangle = \text{sign}(-\widehat{\mathbf{g}}^\top \widehat{\mathbf{v}})\langle \widehat{\mathbf{g}}, \widehat{\mathbf{v}} \rangle \leq 0.$$

Then we are going to upper bound $\langle \nabla f(\mathbf{x}), \widetilde{\mathbf{v}} \rangle$, i.e., with probability at least $1 - \delta'$

$$\begin{aligned}
\langle \nabla f(\mathbf{x}), \widetilde{\mathbf{v}} \rangle &= \langle \widehat{\mathbf{g}}, \widetilde{\mathbf{v}} \rangle + \langle \nabla f(\mathbf{x}) - \widehat{\mathbf{g}}, \widetilde{\mathbf{v}} \rangle \\
&\leq \|\nabla f(\mathbf{x}) - \widehat{\mathbf{g}}\|_2 \|\widetilde{\mathbf{v}}\|_2 \\
&\leq c\epsilon,
\end{aligned}$$

where the last inequality follows from the concentration inequality in Lemma 5.4. Then it follows that

$$\begin{aligned}
f(\mathbf{y}) &\leq f(\mathbf{x}) + \eta \langle \nabla f(\mathbf{x}), \widetilde{\mathbf{v}} \rangle + \frac{\eta}{2} \langle \widetilde{\mathbf{v}}, \nabla^2 f(\mathbf{x}) \widetilde{\mathbf{v}} \rangle + \frac{\rho \eta^3}{6} \|\widetilde{\mathbf{v}}\|_2^3 \\
&\leq f(\mathbf{x}_t) + c\eta\epsilon - \Big(\frac{\epsilon_H \eta^2}{4} - \frac{\rho \eta^3}{6}\Big).
\end{aligned}$$

We first set $c\eta\epsilon \leq \rho\eta^3/6$, which means that $\eta \geq \sqrt{6c\epsilon/\rho}$, then we could guarantee that

$$f(\mathbf{y}) \leq f(\mathbf{x}) - \Big(\frac{\epsilon_H \eta^2}{4} - \frac{\rho \eta^3}{3}\Big).$$

If we take $\eta$ such that

$$\sqrt{\frac{6c\epsilon}{\rho}} \leq \eta < \frac{3\epsilon_H}{4\rho}, \tag{B.1}$$

then the above inequality holds since $c \ll 1$. If we take $\eta = C_H (\epsilon_H/\rho)$, which means that

$$\sqrt{\frac{6c\rho\epsilon}{\epsilon_H^2}} \leq C_H \leq \frac{3}{4},$$

then we can get

$$f(\mathbf{y}) - f(\mathbf{x}) \leq -\Big(\frac{C_H^2}{4} - \frac{C_H^3}{3}\Big)\frac{\epsilon_H^3}{\rho^2} < 0,$$

which guarantees that there will be a sufficient decrease after performing negative curvature step (4.2).

Secondly, we prove that the norm of the gradient increases after the negative curvature step, i.e., $\|\nabla f(\mathbf{x}_{t+1})\|_2 > \epsilon$. According to the previous analysis in Lemma 4.3, we can get the lower bound of $\|\nabla f(\mathbf{x}_{t+1}) - \nabla f(\mathbf{x}_t)\|_2$ as

$$\|\nabla f(\mathbf{y}) - \nabla f(\mathbf{x})\|_2 \geq \eta |\langle \widetilde{\mathbf{v}}, \nabla^2 f(\mathbf{x}) \widetilde{\mathbf{v}} \rangle| - \frac{1}{2}\rho\eta^2 \geq \frac{1}{2}\epsilon_H \eta - \frac{1}{2}\rho\eta^2. \tag{B.2}$$

Then it can be derived that

$$\|\nabla f(\mathbf{y})\|_2 \geq \frac{1}{2}\epsilon_H \eta - \frac{1}{2}\rho\eta^2 - \|\nabla f(\mathbf{x}_t)\|_2 \geq \frac{1}{2}\epsilon_H \eta - \frac{1}{2}\rho\eta^2 - \epsilon,$$

where the last inequality lies in the fact that $\|\nabla f(\mathbf{x}_t)\|_2 \leq \epsilon$ according to the subsampled gradient $\|\widehat{\mathbf{g}}\|_2 \leq \epsilon/2$. In order to guarantee that $\|f(\mathbf{y})\|_2 > \epsilon$, we need to set $\eta$ satisfying

$$\frac{1}{2}\epsilon_H \eta - \frac{1}{2}\rho \eta^2 - \epsilon > \epsilon.$$

In order to guarantee that there exists a solution to the above inequality, we further need to set $\epsilon < \epsilon_H^2/16\rho$. Then similar to the previous analysis in Lemma 4.3, we can get

$$\frac{1}{2} - \frac{1}{2}\sqrt{1 - \frac{16\rho\epsilon}{\epsilon_H^2}} < C_H < \frac{1}{2} + \frac{1}{2}\sqrt{1 - \frac{16\rho\epsilon}{\epsilon_H^2}}. \tag{B.3}$$

Suppose $c$ is sufficiently small, then we could guarantee that if $C_H$ satisfies condition (B.1), it also satisfies condition (B.3). Similar to Lemma 1, we could set the step size as $\eta = \epsilon_H/2c_1\rho$ with $c_1 \geq 1$, and then arrive at the the statements in Lemma 5.5. □

## B.2 Proof of Theorem 5.7

*Proof.* Let $\mathcal{N} = \{k|\|\mathbf{g}_k\|_2 \leq \epsilon/2\}$ and $\mathcal{S} = \{k|\|\mathbf{g}_k\|_2 > \epsilon/2\}$ be two sets, it can be seen that $\mathcal{N}$ contains iterations performing One-Step-Online function and $\mathcal{S}$ consists of iterations running one-epoch SCSG algorithm. According to Lemmas 5.6 and 4.3, we know that at the $k$-th outer loop of Algorithm 4, it either follows that

$$\frac{1}{CL(b/B)^{1/3}}\mathbb{E}\Big[\|\nabla f(\mathbf{x}_k)\|_2^2\Big] \leq \mathbb{E}\big[f(\mathbf{x}_{k-1}) - f(\mathbf{x}_k)\big] + \frac{6 \cdot \mathbb{1}\{B < n\}}{CL(b/B)^{1/3}B} \cdot \mathcal{H}^*, \tag{B.4}$$

or

$$\frac{C'\epsilon_H^3}{\rho^2} \leq \mathbb{E}[f(\mathbf{x}_{k-1}) - f(\mathbf{x}_k)]. \tag{B.5}$$

In fact, Lemma 4.3 only suggests that there exists $\widetilde{C}'$ such that $\widetilde{C}'\epsilon_H^3/\rho^2 \leq f(\mathbf{x}_{k-1}) - f(\mathbf{x}_k)$ holds with probability $1-\delta$. However, we are able to ensure that this inequality holds in expectation as in (B.5). According to Assumptions 3.1 and 3.2 and suppose the step size $\eta < 1$ in Algorithm 3, we know that $f(\mathbf{x}_{k-1}) - f(\mathbf{x}_k) \geq -\rho - L$. Then we have

$$\mathbb{E}[f(\mathbf{x}_{k-1}) - f(\mathbf{x}_k)] \geq \frac{\widetilde{C}'\epsilon_H^3}{\rho^2}(1-\delta) - (\rho+L)\delta.$$

Considering sufficiently small $\delta$ such that $\delta/(1-\delta) \leq \widetilde{C}'\epsilon_H^3/\big(2\rho^2(\rho+L)\big)$, the inequality (B.5) holds with $C' = \widetilde{C}'/2$.

Combining (B.4) and (B.5), performing the summation on both sides over $k$, we have

$$\frac{1}{CL(b/B)^{1/3}}\sum_{k\in\mathcal{S}}\mathbb{E}\Big[\|\nabla f(\mathbf{x}_k)\|_2^2\Big] + \sum_{k\in\mathcal{N}}\frac{C'\epsilon_H^3}{\rho^2} \leq \Delta_f + \sum_{k\in\mathcal{S}}\frac{6\mathcal{H}^*}{CL(b/B)^{1/3}B}.$$

Let $S = |\mathcal{S}|$ and $N = |\mathcal{N}|$, the above formula becomes

$$\frac{1}{CL(b/B)^{1/3}} \sum_{k \in \mathcal{S}} \mathbb{E}\Big[\|\nabla f(\mathbf{x}_k)\|_2^2\Big] + \frac{NC'\epsilon_H^3}{\rho^2} \leq \Delta_f + \frac{6S\mathcal{H}^*}{CL(b/B)^{1/3}B}. \tag{B.6}$$

We then consider $\mathcal{S}_1 = \{k \in \mathcal{S} | \|\mathbf{g}_{k+1}\|_2 > \epsilon/2\}$, and $\mathcal{S}_2 = \{k \in \mathcal{S} | \|\mathbf{g}_{k+1}\|_2 \leq \epsilon/2\}$, whose cardinalities are $S_1$ and $S_2$, respectively. It should be noted that for each $k \in \mathcal{S}_2$, we must have $k+1 \in \mathcal{N}$, which yields $S_2 \leq N + 1$. Then (B.6) can be rewritten as

$$\frac{1}{CL(b/B)^{1/3}} \sum_{k \in \mathcal{S}_1} \mathbb{E}\Big[\|\nabla f(\mathbf{x}_k)\|_2^2\Big] + \frac{1}{CL(b/B)^{1/3}} \sum_{k \in \mathcal{S}_2} \mathbb{E}\Big[\|\nabla f(\mathbf{x}_k)\|_2^2\Big]$$
$$+ N\left(\frac{C'\epsilon_H^3}{\rho^2} - \frac{6\mathcal{H}^*}{CL(b/B)^{1/3}B}\right) \leq \Delta_f + \frac{6(S_1+1)\mathcal{H}^*}{CL(b/B)^{1/3}B}.$$

Then we set the batch size $B$ and mini-batch size $b$ satisfy the following condition

$$B\left(\frac{b}{B}\right)^{1/3} > \frac{12\mathcal{H}^*\rho^2}{CC'L\epsilon_H^3}, \tag{B.7}$$

then all terms on the left-hand side of the above inequality are positive, thus we have

$$\frac{NC'\epsilon_H^3}{2\rho^2} \leq \Delta_f + \frac{6(S_1+1)\mathcal{H}^*}{CL(b/B)^{1/3}B}, \tag{B.8}$$

and

$$\sum_{k \in \mathcal{S}_1} \mathbb{E}\Big[\|\nabla f(\mathbf{x}_k)\|_2^2\Big] \leq CL(b/B)^{1/3}\Delta_f + \frac{6(S_1+1)\mathcal{H}^*}{B}.$$

According to Lemma 5.4 and setting $c = 1/4$, $B = \widetilde{O}(1/\epsilon^2)$, we know that at the $k$-th outer loop of Algorithm 4, for each $k \in S_1$, the inequality $\|\nabla f(\mathbf{x}_k)\|_2 \geq \|\mathbf{g}_k\|_2 - c\epsilon \geq \epsilon/4$ holds with probability at least $1 - \delta_1$. Similarly, for $k \in S_2$, the inequality $\|\nabla f(\mathbf{x}_k)\|_2 \leq \|\mathbf{g}_k\|_2 + c\epsilon \leq \epsilon$ also holds with probability at least $1 - \delta_1$. Then by Markov inequality, we can derive that

$$\sum_{k \in \mathcal{S}_1} \|\nabla f(\mathbf{x}_k)\|_2^2 \leq 2 \sum_{k \in \mathcal{S}_1} \mathbb{E}\Big[\|\nabla f(\mathbf{x}_k)\|_2^2\Big] \leq 2CL(b/B)^{1/3}\Delta_f + \frac{12(S_1+1)\mathcal{H}^*}{B} \tag{B.9}$$

holds with probability at least $1/2$. Thus with probability at least $1/2$, the following holds

$$\frac{S_1\epsilon^2}{4} \leq 2CL(b/B)^{1/3}\Delta_f + \frac{12(S_1+1)\mathcal{H}^*}{B}. \tag{B.10}$$

Assuming $B \geq 96\mathcal{H}^*/\epsilon^2$, we have

$$S_1 \leq \frac{16CL(b/B)^{1/3}\Delta_f + 96\mathcal{H}^*/B}{\epsilon^2}.$$

In the following we are going to upper bound the number of iterations in $\mathcal{N}$. Note that we have already obtained the upper bound of $S_1$, thus the following can be derived from (B.8),

$$\frac{NC'\epsilon_H^3}{2\rho^2} \leq \Delta_f + \frac{96\Delta_f \mathcal{H}^*}{B\epsilon^2} + \frac{576\mathcal{H}^{*2}}{CL(b/B)^{1/3}B^2\epsilon^2} + \frac{6\mathcal{H}^*}{CL(b/B)^{1/3}B}$$
$$\leq 2\Delta_f + \frac{12\mathcal{H}^*}{CL(b/B)^{1/3}B},$$

where the second inequality follows from the assumption $B \geq 96\mathcal{H}^*/\epsilon^2$. Then it can be seen that

$$N \leq \frac{4\Delta_f \rho^2}{C'\epsilon_H^3} + \frac{24\rho^2 \mathcal{H}^*}{C'CL(b/B)^{1/3}B\epsilon_H^3}.$$

Note that we require $B = \widetilde{O}(1/\epsilon^2)$ and $B \geq 96\mathcal{H}^*/\epsilon^2$, which implies that $B = \widetilde{O}(\mathcal{H}^*/\epsilon^2)$. Together with (B.7), we have

$$S_1 = O\left(\frac{L\Delta_f(b/B)^{1/3}}{\epsilon^2} + \frac{1}{B\epsilon^2}\right) = O\left(\frac{L\Delta_f}{\epsilon^{4/3}\mathcal{H}^{*1/3}} + \frac{\rho^2 \Delta_f}{\epsilon_H^3}\right),$$
$$N = O\left(\frac{\rho^2 \Delta_f}{\epsilon_H^3} + \frac{\rho^2}{(b/B)^{1/3}B\epsilon_H^3}\right) = O\left(\frac{\rho^2 \Delta_f}{\epsilon_H^3}\right),$$

where the first equality follows from the fact $b^{1/3} = \max\{1, H^*\rho^2\epsilon^{3/4}/(L\epsilon_H^3)\} \leq 1 + H^*\rho^2\epsilon^{3/4}/(L\epsilon_H^3)$.

By Lemmas 5.2, we know that Algorithm 3 requires $\widetilde{O}(L^2/\epsilon_H^2)$ inner iterations. Moreover, since our algorithm can guarantee escaping from saddle points in one step, thus the number of calls to this function is also upper bounded by $N_\epsilon$, which yields $N = \widetilde{O}(\min\{\rho^2 \Delta_f/\epsilon_H^3, N_\epsilon\})$. In addition, since $T_k$ follows a geometric distribution with mean $B/b$, it can be seen that $T_k \leq O(\log \delta_2(B/b))$ with probability $1 - \delta_2$. Thus, the complexity of one-epoch SCSG is in the order of $O(B \log q) = \widetilde{O}(\mathcal{H}^*/\epsilon^2)$. Now, we can obtain the total runtime complexity of Algorithm 4,

$$T = (S_1 + S_2) \cdot \widetilde{O}\left(\frac{\mathcal{H}^*}{\epsilon^2}\right)\mathbb{T}_g + N \cdot \widetilde{O}\left(\frac{L^2}{\epsilon_H^2}\right)\mathbb{T}_g$$
$$= O\left(\frac{L\Delta_f}{\epsilon^{4/3}\mathcal{H}^{*1/3}} + \frac{\rho^2 \Delta_f}{\epsilon_H^3}\right) \cdot \widetilde{O}\left(\frac{\mathcal{H}^*}{\epsilon^2}\right)\mathbb{T}_g + O\left(\min\left\{\frac{\rho^2 \Delta_f}{\epsilon_H^3}, N_\epsilon\right\}\right) \cdot \widetilde{O}\left(\frac{\mathcal{H}^*}{\epsilon^2} + \frac{L^2}{\epsilon_H^2}\right)\mathbb{T}_g$$
$$= \widetilde{O}\left(\left[\frac{L\Delta_f \mathcal{H}^{*2/3}}{\epsilon^{10/3}} + \frac{\rho^2 \Delta_f \mathcal{H}^*}{\epsilon_H^3 \epsilon^2} + \left(\frac{\mathcal{H}^*}{\epsilon^2} + \frac{L^2}{\epsilon_H^2}\right)\min\left\{\frac{\rho^2 \Delta_f}{\epsilon_H^3}, N_\epsilon\right\}\right]\mathbb{T}_g\right),$$

where the second equality follows from the fact that $S_2 \leq N + 1$. The last thing is to investigate the success probability of Algorithm 4. Note that the one-epoch SCSG function succeeds with probability $(1 - \delta_1)(1 - \delta_2)$, and One-Step-Online succeeds with probability $1 - \delta_1$. Together with the probability introduced by Markov inequality in B.9, the total success probability is $(1 - \delta_1)^S(1 - \delta_2)^S(1 - \delta_1)^N/2$. Note that the probability $\delta_1$ and $\delta_2$ only exist in the logarithmic terms, which can be hided in the notation $\widetilde{O}(\cdot)$. Thus, we consider sufficiently small $\delta_1$ and $\delta_2$ such that $(1 - \delta_1)^S(1 - \delta_2)^S(1 - \delta_1)^N/2 \leq 1/3$, which completes the proof. $\square$

# C Proofs for Finite-Sum Nonconvex Optimization

## C.1 Proof of Theorem 6.5

*Proof.* Let $\mathcal{N} = \{k|\|\mathbf{g}_k\|_2 \leq \epsilon\}$ and $\mathcal{S} = \{k|\|\mathbf{g}_k\|_2 > \epsilon\}$ be two sets for iterations performing One-Step-FiniteSum functions and one-epoch SCSG algorithms, respectively. According to Lemmas 6.4 and 4.3, we know that at the $k$-th outer loop of Algorithm 6, it either follows that

$$\frac{1}{CL(b/B)^{1/3}}\mathbb{E}\Big[\|\nabla f(\mathbf{x}_k)\|_2^2\Big] \leq \mathbb{E}\big[f(\mathbf{x}_{k-1}) - f(\mathbf{x}_k)\big],$$

or

$$\frac{C'\epsilon_H^3}{\rho^2} \leq \mathbb{E}[f(\mathbf{x}_{k-1}) - f(\mathbf{x}_k)].$$

Combining these two cases and perform the summation on both sides over $k$, we have

$$\frac{1}{CL(b/B)^{1/3}}\sum_{k\in\mathcal{S}}\mathbb{E}\Big[\|\nabla f(\mathbf{x}_k)\|_2^2\Big] + \frac{NC'\epsilon_H^3}{\rho^2} \leq \Delta_f. \tag{C.1}$$

We then consider $\mathcal{S}_1 = \{k \in \mathcal{S}|\|\mathbf{g}_{k+1}\|_2 > \epsilon\}$, and $\mathcal{S}_2 = \{k \in \mathcal{S}|\|\mathbf{g}_{k+1}\|_2 \leq \epsilon\}$, whose cardinalities are $S_1$ and $S_2$, respectively. Similar to the proof for Theorem 5.7, we have $S_2 \leq N + 1$. Then the following two inequalities hold,

$$\frac{NC'\epsilon_H^3}{2\rho^2} \leq \Delta_f, \quad \text{and} \quad \sum_{k\in\mathcal{S}_1}\mathbb{E}\Big[\|\nabla f(\mathbf{x}_k)\|_2^2\Big] \leq CL(b/B)^{1/3}\Delta_f. \tag{C.2}$$

By Markov inequality, we have

$$\sum_{k\in\mathcal{S}_1}\|\nabla f(\mathbf{x}_k)\|_2^2 \leq 2\sum_{k\in\mathcal{S}_1}\mathbb{E}\Big[\|\nabla f(\mathbf{x}_k)\|_2^2\Big] \leq 2CL(b/B)^{1/3}\Delta_f$$

holds with probability at least $1/2$. Thus by setting $B = n$ and $b = 1$, we can derive that

$$S_1 \leq \frac{2CL(b/B)^{1/3}\Delta_f}{\epsilon^2} = O\bigg(\frac{L\Delta_f}{\epsilon^2 n^{1/3}}\bigg).$$

From (B.8), the upper bound of $N$ can be directly shown as

$$N \leq \frac{2\rho^2\Delta_f}{C'\epsilon_H^3} = O\bigg(\frac{\rho^2\Delta_f}{\epsilon_H^3}\bigg).$$

By Lemmas 6.1 and 6.2, we know that One-Step-FiniteSum requires $\widetilde{O}(n + n^{3/4}\sqrt{L/\epsilon_H})$ inner iterations and $N = \widetilde{O}\big(\min\{\rho^2\Delta_f/\epsilon_H^3, N_\epsilon\}\big)$. In addition, the one-epoch SCSG epoch has the computational complexity $\widetilde{O}(B) = \widetilde{O}(n)$. Hence, the runtime complexity of Algorithm 6 can be obtained as follows,

$$T = (S_1 + S_2) \cdot \widetilde{O}(n)\mathbb{T}_g + N \cdot \widetilde{O}\Big(n + \frac{n^{3/4}L^{1/2}}{\epsilon_H^{1/2}}\Big)\mathbb{T}_g$$
$$= O\Big(\frac{L\Delta_f}{\epsilon^2 n^{1/3}} + N\Big) \cdot \widetilde{O}(n)\mathbb{T}_g + N \cdot \widetilde{O}\Big(n + \frac{n^{3/4}L^{1/2}}{\epsilon_H^{1/2}}\Big)\mathbb{T}_g$$
$$= \widetilde{O}\Big(\Big[\frac{L\Delta_f n^{2/3}}{\epsilon^2} + \Big(n + \frac{n^{3/4}L^{1/2}}{\epsilon_H^{1/2}}\Big)\min\Big\{\frac{\rho^2\Delta_f}{\epsilon_H^3}, N_\epsilon\Big\}\Big]\mathbb{T}_g\Big),$$

where the second equality follows from the fact that $S_2 \leq N + 1$. Then we are going to analyze the success probability of Algorithm 6. Note that the one-epoch SVRG function succeeds with probability $1 - \delta_2$, and One-Step-FiniteSum succeeds with probability $1 - \delta_1$. Together with the probability introduced by Markov inequality, the total success probability is $(1 - \delta_2)^S(1 - \delta_1)^N/2$. Similarly, considering sufficiently small $\delta_1$ and $\delta_2$ such that $(1 - \delta_2)^S(1 - \delta_1)^N/2 \leq 1/3$, then we are able to complete the proof. □